\documentclass[fleqn,10pt]{wlscirep}
\usepackage[utf8]{inputenc}
\usepackage[T1]{fontenc}
\usepackage{geometry}
\usepackage{indentfirst}
\usepackage{enumitem}
\usepackage[numbers]{natbib}
\usepackage{url}
\usepackage{amsmath, amssymb, amsfonts, amsthm, mathrsfs}
\usepackage{siunitx}
\usepackage{algorithm}
\usepackage[noend]{algpseudocode}

\usepackage{graphicx} 
\usepackage{subcaption}
\usepackage{caption}
\usepackage{multirow} 
\usepackage{booktabs}
\usepackage{array}
\usepackage{tabularx}
\definecolor{user1color}{RGB}{0, 102, 204}
\definecolor{user2color}{RGB}{0, 128, 128}
\usepackage{longtable}

\usepackage{longtable}
\usepackage{makecell}
\usepackage[table]{xcolor}

\renewcommand{\arraystretch}{1.30}

\newcolumntype{L}[1]{>{\raggedright\arraybackslash}p{#1}}
\newcolumntype{C}[1]{>{\centering\arraybackslash}p{#1}}

\definecolor{headerblue}{RGB}{41,65,106}
\definecolor{rowgray}{RGB}{240,243,248}

\usepackage{hyperref}
\usepackage{nameref}
\usepackage[capitalise]{cleveref}

\usepackage{xcolor}
\definecolor{darkblue}{rgb}{0,0,0.7}
\hypersetup{
    colorlinks=true,
    linkcolor=darkblue,
    citecolor=darkblue,
    urlcolor=darkblue
}

\usepackage{listings}
\usepackage{textcomp}

\lstdefinestyle{mystyle}{
    basicstyle=\ttfamily\footnotesize,
    breakatwhitespace=false,
    breaklines=true,
    captionpos=b,
    keepspaces=true,
    numbers=left,
    numbersep=5pt,
    showspaces=false,
    showstringspaces=false,
    showtabs=false,
    tabsize=2
}
\usepackage{comment}
\usepackage{graphicx}
\graphicspath{{figures/}}
\usepackage{gensymb}

\usepackage{xspace}
\newcommand{\dataset}{SomBench\xspace}

\definecolor{lightgray}{gray}{0.9}
\definecolor{codegray}{rgb}{0.5,0.5,0.5}
\definecolor{codepurple}{rgb}{0.58,0,0.82}
\definecolor{backcolour}{rgb}{0.95,0.95,0.92}

\usepackage{soul}
\sethlcolor{yellow}

\title{\dataset: Benchmark Dataset for Advancing Machine Learning in Lunar Science}

\author[1,$\dag$]{Himanshu Patil}
\author[2,$\dag$]{Gabby Nyirjesy}
\author[3,$\dag$]{Rachel A. Slank}
\author[1,$\dag$]{Vishal Gaur}
\author[2,$\dag$]{Daniela Szwarcman}
\author[2,$\dag$]{Paolo Fraccaro}
\author[1,$\dag$]{Nikolaos Dionelis}
\author[4,*,$\dag$]{Michael K. Barker}
\author[5,6,$\dag$]{Andrew Annex}
\author[4,7,$\dag$]{Vishnu Viswanathan}
\author[4,8,$\dag$]{Zachary Morse}
\author[5,$\dag$]{Ethan I. Schaefer}
\author[2]{Hiyam Debary}
\author[1]{Ankur Kumar}
\author[1]{Rohit Lal}
\author[2]{Geoffrey Dawson}
\author[2]{Campbell Watson}
\author[9]{Rebekah I. Dawson-Rigas}
\author[10]{Manil Maskey}
\author[2]{Juan Bernab\'e-Moreno}
\author[10]{Rahul Ramachandran}
\author[1,10,*]{Sujit Roy}

\affil[1]{The University of Alabama in Huntsville, Huntsville, AL, USA}
\affil[2]{IBM Research, Yorktown Heights, NY, USA}
\affil[3]{Science and Technology Institute, Universities Space Research Association (USRA), Huntsville, AL, USA}
\affil[4]{NASA Goddard Space Flight Center, Greenbelt, MD, USA}
\affil[5]{SETI Institute, Mountain View, CA, USA}
\affil[6]{NASA Ames Research Center, Moffett Field, CA, USA}
\affil[7]{University of Maryland, Baltimore County (UMBC), Baltimore, MD, USA}
\affil[8]{Howard University, Washington DC, USA}
\affil[9]{NASA Headquarters, Washington, DC, USA}
\affil[10]{NASA Marshall Space Flight Center, Huntsville, AL, USA}
\affil[*]{corresponding author: Sujit Roy (sujit.roy@nasa.gov), Michael K. Barker (michael.k.barker@nasa.gov)}
\affil[$\dag$]{Equal Contribution to this work and share first-authorship. Names were ordered randomly.}

\begin{abstract}
Lunar orbital missions, such as Lunar Reconnaissance Orbiter, Kaguya/SELENE, Gravity Recovery and Interior Laboratory, and Lunar Prospector, among others, provide rich multi-instrument observations, but their heterogeneity in sampling, projection, and conventions limits reproducible machine learning (ML). We introduce \dataset, a unified, spatially-aligned, ML-ready lunar dataset aggregating 30+ co-registered layers from ten instruments across four missions, spanning \SI{1}{\meter} to \SI{20}{\kilo\meter}/pixel and covering \textpm82\degree{} latitude in 90 Lunar Transverse Mercator zones with two polar stereographic caps. 
An image-anchored tiling pipeline yields pretraining-ready multimodal tile views with leakage-safe splits, distributed as netCDF with Parquet catalogs. An application benchmark suite spans impact processes, volcanic history, and polar volatiles. Baseline experiments with ResNet-50 and SwinV2-B models confirm that each benchmark task is learnable from the released inputs, establishing reference points for future model development.

\end{abstract}

\begin{document}  

\flushbottom
\maketitle
\section{Background and Summary}

Lunar science and exploration have entered a data-rich era in which long duration orbital campaigns provide sustained, multi-instrument measurements of surface morphology, composition, topography, thermophysical state, and polar illumination environments. Over almost two decades, NASA's Lunar Reconnaissance Orbiter (LRO) alone has generated nearly 2~PB of data, including broadband meter-scale and multi-band hundred-meter-scale imagery \cite{robinson2010lunar}, laser altimetry \cite{smith2010lunar}, thermal emission \cite{paige2010lunar}, and radar observations \cite{nozette2010lunar}. JAXA's Kaguya/SELENE mission added high-resolution multispectral imaging and derived mineralogy \cite{lemelin2015lunar,lemelin2016global,lemelin2019compositions} and stereo topography \cite{barker2016new}. NASA's Gravity Recovery and Interior Laboratory (GRAIL) mission mapped the lunar gravity field at unprecedented resolution \cite{zuber2013gravity,konopliv2014high}, and Lunar Prospector contributed neutron-derived constraints on near-surface hydrogen \cite{feldman1999lunar,feldman2001evidence}. Together, these observations enable local to global investigations of key lunar processes, including impact cratering and regolith evolution, volcanic activity and modification, and the delivery and mobility of volatiles in permanently shadowed regions (PSRs), while also supporting operational workflows such as landing site characterization and hazard assessment. At the same time, the breadth and heterogeneity of modern lunar datasets make it increasingly difficult to build reproducible analysis pipelines and to compare results across studies, instruments, and observing geometries.
\\
Machine learning (ML) offers a complementary pathway for extracting scalable, consistent information from these large datasets. However, successful ML development in lunar science is often constrained by practical barriers. Different instruments have different spatial sampling, map projections, and coverage. Observing geometry also widely varies and can dominate apparent surface contrast, while many commonly used derived products are distributed in forms that are not easily interoperable. As a result, even when strong task-specific models exist, it can be challenging (i) to train them on consistent inputs, (ii) to evaluate them against standardized splits and protocols, and (iii) to compare methods fairly across research groups and across lunar science problems.

Despite this wealth of data, no unified, co-registered dataset exists that brings these heterogeneous observations into a common spatial framework suitable for modern machine learning. Large-scale, multi-modal pretraining corpora have driven major advances in Earth observation~\cite{cong2022satmae, stewart2023ssl4eo, fuller2023croma, reed2023scale}, and curated ML-ready benchmarks have played an analogous role in other science domains such as heliophysics \citep{roy2025suryabench}, but no comparable resource exists for planetary science. Existing lunar ML efforts have instead relied on single-instrument products or hand-curated regions of interest, which limits direct comparison across studies and constrains broader adoption by both lunar scientists and ML practitioners. 

To address these challenges, we introduce \dataset, a curated, publicly accessible, ML–ready benchmark dataset for lunar science, designed explicitly to support both self-supervised foundation model pretraining and standardized downstream evaluation. \dataset aggregates key lunar observations and widely used derived map products into standardized, geodetically consistent inputs, with harmonized metadata, uniform conventions, and AI-ready formats intended to reduce domain-specific overhead for both lunar scientists and ML practitioners. It also provides a consistent and reusable testing environment that enables systematic development and evaluation of ML methods on lunar data. Recent efforts have similarily begun to develop multimodal datasets for machine learning. For example, Moonstone \cite{prasad2026moonstonemultimodalfoundationmodel} includes 28 channels and 281 GB of data at a fixed resolution of ~237 m/pixel. \dataset extends this emerging capability by providing 32 channels and ~42.5 TB of data, more than 150 times the data volume, while spanning global-scale observations to meter-scale NAC imagery. This combination of data volume and spatial-resolution range supports both broad regional analysis and fine-scale investigations of lunar morphology within a common dataset framework. 

\dataset provides two complementary components. First, it includes a core multi-instrument lunar dataset that serves as a pre-training corpus, co-registering 32 data layers from ten instruments across four missions (LRO, Kaguya/SELENE, GRAIL, and Lunar Prospector, described in Methods) that span resolutions from \SI{1}{\meter} to \SI{20}{\kilo\meter}/pixel. The modalities also range from visible/near-infrared imagery through normalized reflectance and mineralogy derivatives, topography and topographic derivatives, radar scattering, and thermophysical and illumination-relevant products to gravity context. An adaptive, image-anchored tiling pipeline assembles these layers into multimodal tile stacks along two parallel tracks. The first is a low-resolution track anchored to Wide Angle Camera (WAC) images (\SI{51.2}{\kilo\meter} tiles at \SI{100}{\meter}/pixel) and the other is a high-resolution track anchored to Narrow Angle Camera (NAC) images (\SI{512}{\meter} tiles at \SI{1}{\meter}/pixel). This enables multi-scale learning within a single framework while handling lunar-specific projection challenges (Lunar Transverse Mercator zones, polar stereographic caps, antimeridian discontinuities). The corpus is distributed as compressed netCDF tiles with embedded georeference metadata, Parquet-based catalogs, a PyTorch \texttt{Dataset} class, per-layer quality gating, and geographic train/validation/test splits that prevent spatial data leakage. Because benchmark validity depends on reproducibility, all layers are released as map-projected products with uniform conventions for projection, units, and metadata, with separate global and polar stereographic variants where appropriate, including polar-focused layers (illumination and PSR descriptors, thermophysical indicators relevant to cold trap stability, and composition-sensitive products) that remain diagnostic under the challenging lighting conditions near the poles.

Second, \dataset defines a set of application benchmark datasets (hereafter “benchmarks”) that target recurring lunar science and exploration tasks. These benchmarks are packaged with standardized dataset splits, evaluation metrics, and baseline implementations to enable direct comparison across methods and to facilitate rapid iteration on model architectures, training objectives (including self-supervised approaches), and multimodal fusion strategies.

Consistent with common lunar science workflows, our proposed \dataset benchmarks emphasize tasks that combine feature identification, quantitative characterization, contextualization, and classification. This initial benchmark suite focuses on representative problems spanning three broad science themes, impact processes, volcanic history, and polar volatiles, that are both scientifically meaningful and representative of the technical challenges posed by lunar data (multi-scale structure, heterogeneous sampling, and illumination-driven appearance changes). The tasks include: (i) detection and delineation of impact craters and related morphologic units, (ii) segmentation and classification of geologic and volcanic features that are difficult to catalog manually at global scale, and (iii) polar mapping tasks that integrate illumination, thermophysical context, and surface property proxies to support analyses of volatile retention environments. Together, these benchmarks are intended to serve as a shared foundation for reproducible ML evaluation on lunar data and to accelerate progress toward robust, interpretable models suitable for both scientific investigation and operational decision support. \dataset is designed to be usable by both lunar domain experts and ML practitioners, and includes detailed documentation, consistent preprocessing, and rich metadata to support seamless interoperability across instruments and tasks.

The remainder of this paper follows the data from acquisition to application. The Methods section describes the contributing missions and instruments, the raw and derived data products drawn from their archives, the preprocessing that converts raw images into analysis-ready rasters, the construction of the ML-ready pre-training dataset, and the application benchmark datasets built on top of it. We then document the released data records, validate the datasets by training baseline models on every benchmark, and conclude with usage notes and known limitations.

\section{Methods} 

\hspace*{2em}\dataset comprises (i) a core multi-instrument lunar dataset of harmonized global and polar products that serves as the pre-training corpus, and (ii) a set of benchmark task datasets with standardized splits and evaluation protocols. The core dataset provides co-registered, map-projected (i.e., orthorectified) inputs spanning multi-scale orbital imaging, normalized reflectance, topography and terrain derivatives, radar scattering, gravity context, mineral abundance, and thermophysical and illumination-relevant layers. \dataset's core lunar datasets are organized to reflect how lunar investigations are typically performed and how ML models are most effectively trained and evaluated on heterogeneous planetary data. We coupled broad regional context with meter-scale morphology and pair these observations with terrain products that support quantitative measurement and reduce ambiguity introduced by illumination and viewing geometry. This core stack is intended to be usable in two complementary modes: (i) a standardized input corpus for ML representation learning and benchmark evaluation, and (ii) an analysis-ready geospatial context stack for lunar scientists conducting reproducible cross-modal comparisons. This section follows the data from source to ML-ready form. We first introduce the contributing missions and instruments and the archives from which their data are obtained. We then describe the individual imagery, terrain models, and static map products that constitute the core inputs, then we detail the preprocessing that converts raw image products into calibrated, map-projected rasters. Next, we then present the tiling pipeline that assembles these inputs into the pre-training dataset. Finally, we conclude with the application benchmark datasets that operationalize \dataset into reproducible evaluations.

\subsection{Lunar Missions and Instruments}
\label{subsec:missions}

\dataset is built from observations acquired by ten instruments across four lunar orbital missions. These missions were selected because their long-baseline datasets (i) provide global coverage, (ii) include complementary measurement physics, and (iii) are routinely combined in lunar science workflows. Together, they motivate \dataset's core design choices: benchmark inputs must support cross-modal fusion, tolerate heterogeneous spatial sampling, and remain robust to the illumination and geometry effects that are central to lunar data and especially acute at the poles. Table~\ref{tab:data_sources} summarizes the instruments, their native resolutions, bands and products, and the forms in which their data enter \dataset.

\begin{table}[htbp]
\centering
\caption{Missions, instruments, and data products included in \dataset.}
\label{tab:data_sources}
\renewcommand{\arraystretch}{1.4}
\small
\begin{tabularx}{\textwidth}{l l l X X}
\toprule
\textbf{Instrument} & \textbf{Mission} & \textbf{Resolution} & \textbf{Bands / Products} & \textbf{Data Type} \\
\midrule
LROC NAC & LRO & ${\sim}1$\,m & Panchromatic ($400 - 760$ nm) & Individual images, stereo DTMs + related topographic maps \\
\addlinespace
LROC WAC & LRO & 100\,m\,--\,500\,m & VIS: 415, 566, 604, 643, 689\,nm; UV: 321, 360\,nm; morphologic mosaic; multi-band normalized reflectance mosaics; TiO$_2$ abundance & Individual images, global image mosaics \\
\addlinespace
LOLA & LRO & 20\,m\,--\,1\,km & Elevation, slope, aspect, roughness, polar 1064 nm normal albedo, polar illumination, PSRs & Lidar-based global and polar DTMs + related topographic maps; other derived products \\
\addlinespace
Diviner & LRO & 240\,m\,--\,15\,km & $0.35 - 400 \mu \rm m$; rock abundance, $T_{\mathrm{reg}}$ anomaly, H-parameter, $T_{\mathrm{bol}}$ (24 phases), ice stability depth & Thermal emission derived global and polar maps \\
\addlinespace
Mini-RF & LRO & 90\,m & Monostatic S-band radar (12.6 cm) CPR, S1 reflectivity & Controlled, orthorectified, geometrically-normalized global and polar mosaics \\
\addlinespace
TC & Kaguya/SELENE & 60\,m & Panchromatic ($430 - 850$ nm); elevation, slope, aspect & Global stereo DTM mosaic + related topographic maps\\
\addlinespace
MI & Kaguya/SELENE & 60\,m\,--\,1\,km & 415, 750, 900, 950, 1000, 1050, 1250, 1550\,nm; normalized reflectance mosaics + mineralogy (CPX, OPX, olivine, plagioclase, plagioclase grain size, FeO, OMAT, npFe, smFe, mpFe) & Multi-band imaging mosaics + derived mineralogy \\
\addlinespace
SP & Kaguya/SELENE & 1\,km & Olivine, plagioclase, HCP, LCP, FeO, OMAT, npFe & Polar spectral mineralogy \\
\addlinespace
GRAIL & GRAIL & 20\,km & Free-air gravity disturbance & Derived global map \\
\addlinespace
Neutron Spectrometer & Lunar Prospector & 15\,km & Hydrogen abundance & Polar volatile detection \\
\bottomrule
\end{tabularx}
\end{table}

\textit{Lunar Reconnaissance Orbiter (LRO)}

LRO is a NASA mission launched in June 2009 with the primary objective of enabling future human and robotic exploration of the Moon through detailed characterization of the lunar surface and environment \citep{chin2007lunar,vondrak2010lunar}. The spacecraft operates in a near-polar orbit at altitudes between \textasciitilde{}30 and 200 km and, now in extended mission phases, continues to support both scientific discovery and exploration planning \citep{vondrak2010lunar}. LRO provides the backbone of \dataset through five of its instruments. The LRO Camera (LROC) \citep{robinson2010lunar} comprises two Narrow Angle Cameras (NAC), which acquire broadband panchromatic (400--760 nm) images at a typical resolution of 0.5--1.5 m/pixel, and the Wide Angle Camera (WAC), which images in two ultraviolet bands (321 and 360 nm, \textasciitilde{}500 m/pixel) and five visible bands (415, 566, 604, 643, and 689 nm, \textasciitilde{}100 m/pixel). The WAC captures global morphology and multispectral reflectance under controlled photometric normalization, supporting texture- and albedo-driven learning tasks, while the NAC provides detailed monochromatic imagery at fine resolution and, through targeted stereo acquisitions, enables meter-scale topographic reconstruction. The Lunar Orbiter Laser Altimeter (LOLA) provides a globally consistent geodetic reference frame, topography, surface roughness, 1064 nm normal albedo, and modeled illumination products that are essential for terrain-aware interpretation and illumination modeling \cite{smith2010lunar}. The Diviner Lunar Radiometer Experiment (Diviner) measures reflected solar and emitted infrared radiation in nine spectral channels spanning \textasciitilde{}0.35--400 $\mu$m, supplying global thermophysical constraints that encode regolith state and diurnal temperature behavior, particularly relevant for polar environments and volatile stability \cite{paige2010lunar}. Mini-RF is a hybrid-polarimetric synthetic aperture radar operating primarily at S-band (12.6 cm) that adds sensitivity to wavelength-scale roughness and dielectric structure not uniquely constrained by optical data \cite{raney2007hybrid,nozette2010lunar}.

\textit{Kaguya/SELENE}

The Japanese Aerospace Exploration Agency's (JAXA) Kaguya/SELENE mission (2007--2009) \citep{kato2010kaguya} contributes three instruments to \dataset. The Terrain Camera (TC), a panchromatic (430--850 nm) stereo imager, underpins the stereo-derived topography that is merged with LOLA altimetry into a unified global elevation model \cite{barker2016new}. The Multi-band Imager (MI) acquired visible--near-infrared imagery in eight bands between 415 and 1550 nm, from which normalized reflectance mosaics and derived mineralogy and maturity products supporting compositional and maturity-informed learning tasks are produced \cite{ohtake2008performance,lemelin2015lunar,lemelin2016global,lemelin2019compositions}. The Spectral Profiler (SP), a nadir-looking VIS--NIR spectrometer spanning roughly 0.5--2.6 $\mu$m, provides derived mineral and maturity maps that extend compositional constraints into the polar regions \cite{haruyama2008global,yamamoto2014calibration}.

\textit{GRAIL and Lunar Prospector}

The Gravity Recovery and Interior Laboratory (GRAIL) twin-spacecraft mission (2011--2012) mapped the lunar gravity field through Ka-band inter-satellite tracking, yielding free-air gravity products that provide a static geophysical context layer linking surface expressions to subsurface mass structure \cite{zuber2013gravity,konopliv2014high}. The Lunar Prospector Neutron Spectrometer (1998--1999) \citep{binder1998lunar} measured the epithermal neutron suppression associated with near-surface hydrogen, providing hydrogen abundance maps that anchor volatile-related geochemical context at the poles \cite{feldman1999lunar,Lawrence2022}.

\textit{Raw Data Products and Archives}

All source data are publicly archived through NASA's Planetary Data System (PDS) and affiliated mission archives. \dataset ingests two classes of input from these archives. The first is individual raw image products: LROC Experiment Data Records (EDRs), which contain the uncalibrated sensor data for each observation together with label metadata such as viewing and illumination geometry, timing, and data quality flags. We select, radiometrically calibrate, and map-project these products ourselves, as described in Sections~\ref{subsec:imagery} and \ref{subsec:preprocessing}. The second is derived, analysis-ready map products, which includes global and polar mosaics, terrain models, and geophysical and thermophysical maps produced by the instrument teams or independent researchers, which we spatially harmonize onto common grids but do not re-derive. The following subsections describe both classes of input in detail.

\subsection{Core Lunar Datasets}
The core \dataset inputs fall into the two classes introduced above: individual raw imagery and derived products. The individual raw imagery consisting of LROC NAC and WAC EDRs that we select from the PDS archive and then calibrate and map-project ourselves (Section~\ref{subsec:preprocessing}). The derived products, which include NAC stereo DTMs and the global and polar static map suite were produced by the instrument teams or independent researchers and harmonized here onto common grids. We describe each in turn.

\subsubsection{Individual Imagery}
\label{subsec:imagery}

\textit{LROC NAC Imagery}

\begin{figure}[h]
    \centering
    \includegraphics[width=0.8\textwidth]{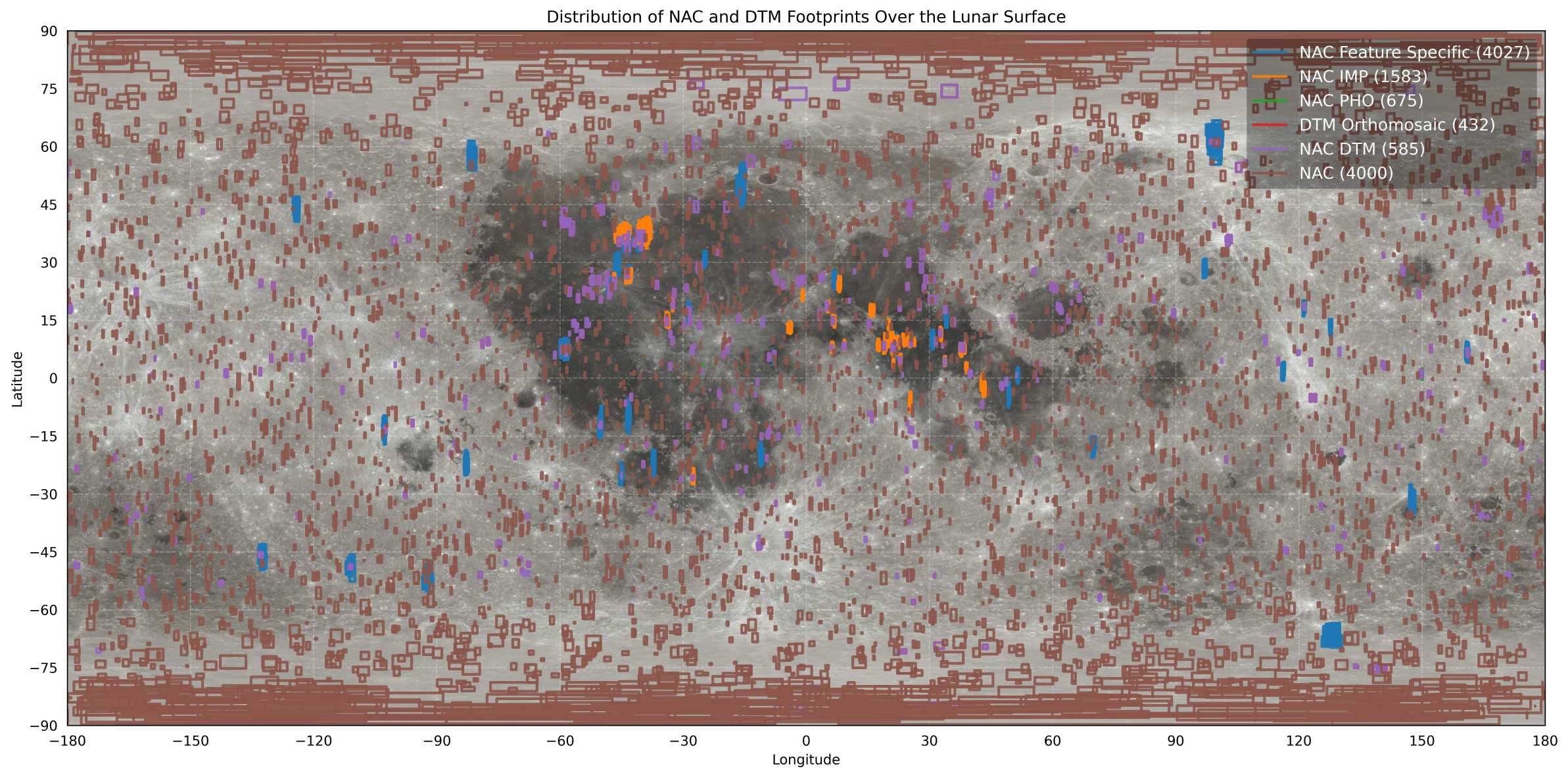}
    \caption{The distribution of NAC, NAC Feature Specific, NAC IMP, NAC PHO, NAC DTMs and DTM Orthomosaic images over the lunar surface in the  \dataset dataset.}
    \label{fig:nac-coverage}
\end{figure}

Over the ongoing LRO mission, the NAC has acquired almost three million images globally at its typical resolution of 0.5–1.5 m/pixel (Section~\ref{subsec:missions}), allowing scientists to study the lunar surface in fine detail.
We select 4,000 images from this huge corpus with varying illumination conditions and locations (labeled "NAC" in Fig.~\ref{fig:nac-coverage}). 
These images were selected in a multi-step approach. First, we pre-filtered each instrument’s image catalog to remove nighttime (INCIDENCE\_ANGLE$>$90) or otherwise suspect images (DATA\_QUALITY\_FLAG$>$0). Then, we binned 4 geometry variables (INCIDENCE\_ANGLE, CENTER\_LATITUDE, CENTER\_LONGITUDE, and SUB\_SOLAR\_GROUND\_AZIMUTH) into discrete categorical bins, and combined those into a single categorical string key that identified each unique joint-bin cell. Finally, after dropping rare joint-bins with fewer than 50 images ($\sim1,000$ images total), we used sklearn.train\_test\_split to draw a stratified sample of exactly 4,000 images.

We also select another set of $\sim4,000$ images that include specific features of high scientific interest (e.g., swirls, lobate scarps, impact melt, pyroclastic deposits, etc., labeled "NAC Feature Specific" in Figure~\ref{fig:nac-coverage}), and another $\sim1,500$ images that contain irregular mare patches (IMPs, labeled "NAC IMP" in Figure~\ref{fig:nac-coverage}).
There are 10 locations on the Moon with many repeat observations intended for detailed photometric studies due to their importance for science and exploration \footnote{An additional two locations were publicly released during the writing of this paper and are not included in \dataset}. These sites, designated "NAC PHO" and listed in Table \ref{tab:img_files}, were imaged under multiple illumination geometries to allow detailed study of their geology and regolith properties. Importantly, the LROC instrument team has publicly released co-registered images for these sites making them ideally suited to ML applications. \dataset contains a total of 675 NAC PHO images shown in Figure~\ref{fig:nac-coverage}. An additional 432 orthomosaics co-registered to stereo topographic models are included and labeled in Figure~\ref{fig:nac-coverage} as "NAC DTM Orthomosaic" (see Sec.~\ref{sec:dtms}). In total, \dataset contains 11,302 NAC images across the entire lunar surface, visible in Figure~\ref{fig:nac-coverage}.

\begin{table}[htbp]
\centering  
\begin{tabular}{l l r}
\hline 
\textbf{Region ID} &\textbf{NAC PHO Site Name} &\textbf{Number of images} \\
\hline
E009S3481 & Apollo 15 SIVB Impact & 83 \\
E010N0230 & Apollo 11 & 99 \\
E011N1499 & Highlands & 88 \\
E018N3346 & Mare Impact Melt & 54 \\
E041N1229 & King Crater Ejecta & 32 \\
E074N3010 & Reiner Gamma & 20 \\
E090S0155 & Apollo 16 & 65 \\
E186N1213 & Highlands Mixed Spectra & 100 \\
E199N0308 & Apollo 17 & 54 \\
E207N3357 & March 17 Impact & 80 \\
\hline
\end{tabular}
\caption{Co-registered NAC Photometry (PHO) locations and number of images.}
\label{tab:img_files}
\end{table}

\textit{LROC WAC Imagery}

\begin{figure}[h]
    \centering
    \includegraphics[width=0.8\textwidth]{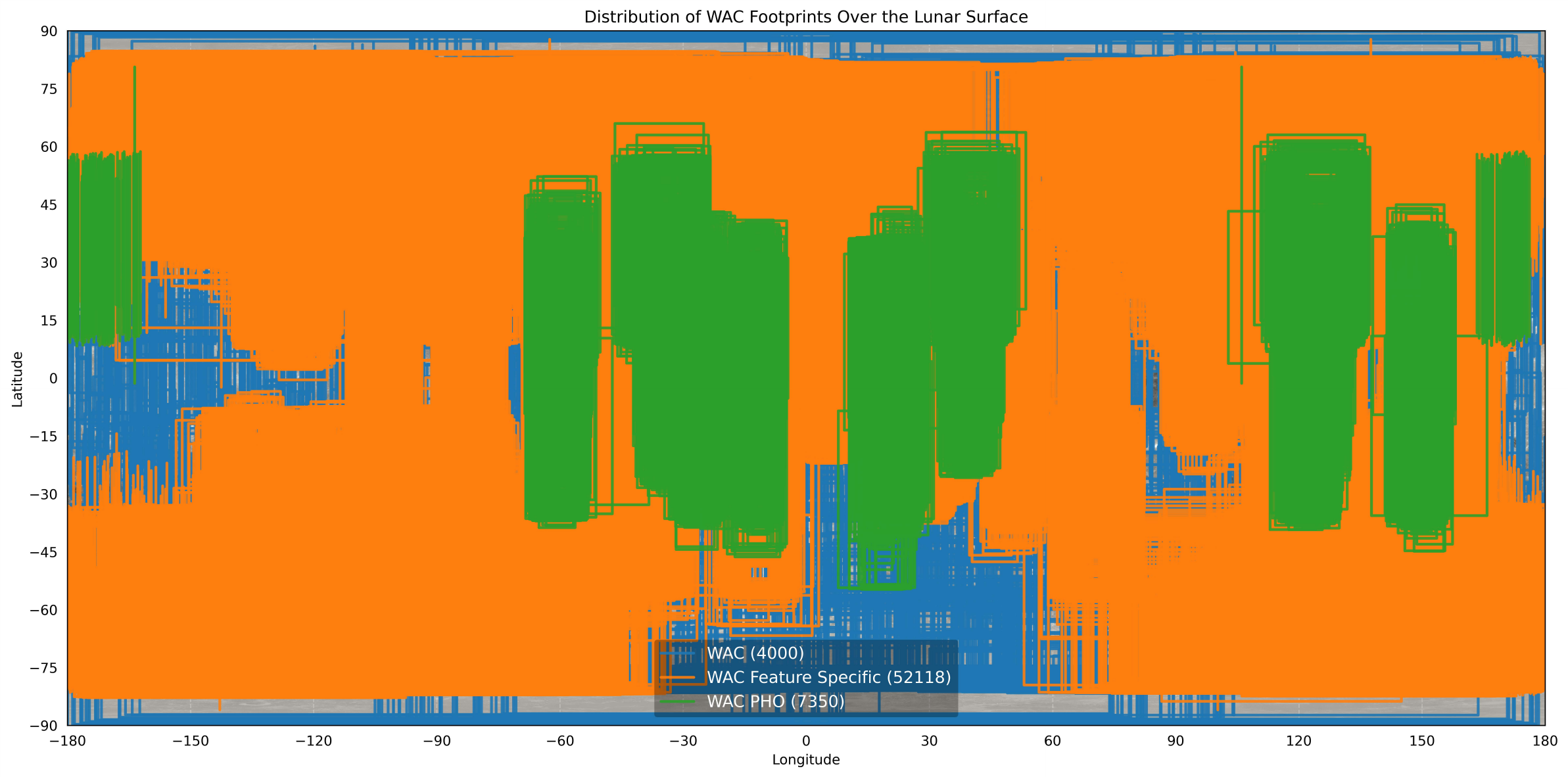}
    \caption{Image presents the distribution of WAC, WAC Feature Specific and WAC PHO Site images used in the \dataset Dataset.}
    \label{fig:wac-coverage}
\end{figure}

The LROC WAC acquires two-band ultraviolet (UV) imagery at a typical resolution of $\sim500$ m/pixel and five-band visible imagery at $\sim100$ m/pixel (Section~\ref{subsec:missions}). Similar to the NAC images, WAC also has a global coverage of the lunar surface with over 1.2 million images. Most of the surface has coverage with multiple images and incidence angles. 
Using the same stratified sampling strategy as for the NAC, we select 4,000 images from the WAC dataset with varying illumination conditions and location (labeled "WAC" in Figure~\ref{fig:wac-coverage}). As with the NAC, another set of 52,110 images were selected to include specific high science value features (labeled "WAC Feature Specific" in Figure~\ref{fig:wac-coverage}). We also include $\sim7,300$ WAC images from the 10 NAC PHO sites (labeled "WAC PHO" in Figure~\ref{fig:wac-coverage}). In total, \dataset contains 63,468 WAC images across the entire lunar surface, visible in Figure~\ref{fig:wac-coverage}.

\subsubsection{NAC Digital Terrain Models (DTMs)} 
\label{sec:dtms}

As part of its normal operations, the NAC instrument acquires stereo pairs to allow the construction of DTMs (digital terrain models, interchangeable with digital elevation models, DEMs, in this paper). These NAC stereo DTMs were produced by the LROC instrument team using stereo photogrammetry, which derives elevation from the geometric differences between two images of the same surface acquired from different viewing angles. In practice, a software correlator identifies matching image patches across the stereo pair and estimates their offset (parallax) in disparity space, which is then converted into topography \cite{beyer2018ames,henriksen2017extracting}. 
The horizontal resolution and vertical precision of the resulting DTM depends on various factors, such as image resolution, spacecraft altitude, stereo convergence angle, terrain texture, and the stereo correlation algorithm \cite{kirk2008ultrahigh,henriksen2017extracting,beyer2018ames}. With typical resolutions of $\sim2-5$ m/pix, the NAC stereo DTM dataset captures fine topographic details unresolved in the global static map products and it samples a diversity of terrains. Thus, it provides valuable information on lunar geology not available with other datasets.
\dataset uses 585 stereo DTMs with corresponding slope and aspect derivatives and the previously mentioned 432 co-registered orthomosaics (the merged map-projected image pairs), shown in Figure ~\ref{fig:nac-coverage}.

\subsubsection{Global Static Map Products}
Global static maps for the lunar surface are standardized, map-projected mosaics created by the instrument teams or independent researchers from data collected by multiple orbiters and instruments. The mosaics were generally created by combining the measurements taken on different orbits under varied illumination conditions. In some cases, such as the WAC Global Morphologic Mosaic, the illumination conditions of the mosaicked data were specifically chosen to minimize illumination differences. These maps show unique geophysical properties of the lunar surface, such as reflectance, geomorphology, mineralogy, topography, terrain roughness, surface and sub-surface rock abundance, gravity, regolith temperature, and thermophysical behavior.
All the different maps are acquired from different missions and satellites. Given the variation between the satellite sensors (e.g., cameras, spectrometers, altimeters), acquisition times, and illumination geometries, etc., the data go through preprocessing and correction steps to bring them onto a common lunar grid before combining them to create the mosaics. The suite includes both global maps and polar-focused products that additionally support analysis of illumination-driven processes and volatile stability near the poles. The polar-focused products are delivered in polar stereographic projection with extents defined relative to the pole. Together, these maps provide comprehensive information about the lunar surface and subsurface, summarized in Table~\ref{tab:static-map-products}.

%
%

\rowcolors{2}{rowgray}{white}

{\footnotesize
\begin{longtable}{
  C{0.35cm}  
  L{2.4cm}   
  L{1.8cm}   
  L{1.85cm}  
  L{1.7cm}   
  L{3.5cm}   
  L{1.1cm}   
}
\caption{Summary of lunar global static map products. A dash indicates a
product released for only one of the two regions; polar latitude ranges
apply at both poles.}
\label{tab:static-map-products}\\
\rowcolor{headerblue}
\textcolor{white}{\textbf{\#}} &
\textcolor{white}{\textbf{Dataset}} &
\textcolor{white}{\textbf{Instrument / Source}} &
\textcolor{white}{\textbf{Global (res.; coverage)}} &
\textcolor{white}{\textbf{Polar (res.; lat., both poles)}} &
\textcolor{white}{\textbf{Key Purpose / Products}} &
\textcolor{white}{\textbf{Ref.}} \\
\endfirsthead
\multicolumn{7}{l}{\small Table~\ref{tab:static-map-products} \textit{(continued)}}\\
\rowcolor{headerblue}
\textcolor{white}{\textbf{\#}} &
\textcolor{white}{\textbf{Dataset}} &
\textcolor{white}{\textbf{Instrument / Source}} &
\textcolor{white}{\textbf{Global (res.; coverage)}} &
\textcolor{white}{\textbf{Polar (res.; lat., both poles)}} &
\textcolor{white}{\textbf{Key Purpose / Products}} &
\textcolor{white}{\textbf{Ref.}} \\
\endhead

\midrule
\multicolumn{7}{c}{\small\itshape Continued on next page\ldots} \\
\endfoot

\bottomrule
\endlastfoot

1 &
LROC WAC Global Morphologic Mosaic &
LRO WAC (643\,nm) &
100\,m/px; 90\textdegree S--90\textdegree N &
100\,m/px; 60--90\textdegree &
Surface morphology emphasis: crater rims, wrinkle ridges, flow fronts, blocky ejecta, topographic shading &
\cite{speyerer2011lunar,robinson2012exploring,wagner2015new,speyerer2013persistently} \\

2 &
LROC WAC Normalized Multi-band Reflectance Mosaics &
LRO WAC (7-band UV--VIS) &
500\,m/px; 60\textdegree S--60\textdegree N &
\textemdash{} &
Photometrically corrected reflectance (I/F); comparative spectral \& compositional analysis; band-ratio work &
\cite{sato2017lunar,boyd2012lunar,sato2014resolved} \\

3 &
LROC WAC Normalized Higher-Resolution Reflectance Mosaic &
LRO WAC (643\,nm) &
100\,m/px; 90\textdegree S--90\textdegree N &
100\,m/px; 60--90\textdegree &
Higher-resolution morphology / registration anchor for the WAC normalized reflectance suite &
\cite{sato2017lunar,sato2014resolved,wagner2015new,speyerer2013persistently} \\

4 &
Kaguya MI Multi-band Normalized Reflectance Mosaics &
Kaguya MI (8-band VIS--NIR) &
60\,m/px; 55\textdegree S--55\textdegree N &
\textemdash{} &
Mafic mineral absorption ($\sim$1\,\textmu m); broadband maturity \& albedo; spectral comparison across terrains &
\cite{lemelin2015lunar,ohtake2008performance,lemelin2016global,lemelin2019compositions} \\

5 &
Kaguya MI Derived Mineralogy &
Derived from MI multi-band reflectance inversion &
60\,m/px, 1\,km/px; 55\textdegree S--55\textdegree N &
\textemdash{} &
\textbf{60\,m/px:} olivine, OPX, CPX, plagioclase, FeO, plag.\ grain size, OMAT. \textbf{1\,km/px:} smFe, mpFe, npFe &
\cite{lemelin2015lunar,lemelin2016global,lemelin2019compositions,Trang2019} \\

6 &
Kaguya SP Derived Mineralogy &
Derived from Kaguya SP hyperspectral inversion &
\textemdash{} &
1\,km/px; 80--90\textdegree &
Olivine, OPX (low-Ca), CPX (high-Ca), plagioclase, FeO, OMAT, npFe &
\cite{haruyama2008global,yamamoto2014calibration} \\

7 &
LROC WAC TiO\textsubscript{2} &
Derived from LRO WAC band ratio (321/415\,nm) &
400\,m/px; 70\textdegree S--70\textdegree N &
\textemdash{} &
TiO\textsubscript{2} abundance (wt\%); mare basalt compositional discrimination; regional stratigraphy &
\cite{sato2017lunar} \\

8 &
Merged / Polar Topography (SLDEM2015 / LOLA DEM) \& Derivatives &
LOLA + Kaguya TC stereo (global); LOLA altimetry (polar) &
60\,m/px; 60\textdegree S--60\textdegree N &
60\,m/px; 60--90\textdegree &
Elevation (DEM); slope (gradient magnitude); aspect (downslope azimuth); aspect sine \& cosine (wraparound-stable) &
\cite{barker2016new,barker2023,barker2025large} \\

9 &
LOLA Roughness &
Derived from LOLA single-shot footprints &
1\,km/px; 90\textdegree S--90\textdegree N &
1\,km/px; 40--90\textdegree &
Surface roughness at a 50-m baseline; regolith development, degradation, geologic age &
\cite{kreslavsky2013lunar,neumann2015} \\

10 &
LOLA Average Illumination &
Derived from LOLA DEMs &
\textemdash{} &
120\,m/px; 75--90\textdegree &
Average solar illumination over a lunar precession cycle ($\sim$18.6\,yr) &
\cite{mazarico2011illumination} \\

11 &
LOLA Permanently Shadowed Regions (PSRs) &
Derived from LOLA DEM &
\textemdash{} &
20\,m/px; 80--90\textdegree &
Map of PSRs: regions never directly illuminated by the Sun &
\cite{barker2023,barker2025large,mazarico2011illumination} \\

12 &
LOLA 1064-nm Normal Albedo &
LOLA laser return energy &
\textemdash{} &
1\,km/px; 50--90\textdegree &
Zero-phase reflectance independent of illumination geometry &
\cite{Lemelin2016} \\

13 &
LRO Mini-RF Radar Reflectivity Mosaic &
LRO Mini-RF hybrid-polarimetric SAR (S-band, 12.6\,cm) &
90\,m/px; 90\textdegree S--90\textdegree N &
90\,m/px; 80--90\textdegree &
Radar backscatter: sensitivity to wavelength-scale surface roughness and dielectric contrasts &
\cite{fassett2024,raney2007hybrid,nozette2010lunar,raney2010lunar,spudis2013evidence} \\

14 &
LRO Mini-RF Circular Polarization Ratio (CPR) Mosaic &
LRO Mini-RF hybrid-polarimetric SAR (S-band, 12.6\,cm) &
90\,m/px; 90\textdegree S--90\textdegree N &
90\,m/px; 80--90\textdegree &
Scattering behavior linked to blockiness, multiple scattering, subsurface structure, and composition (not a standalone ice detector; see caveats) &
\cite{nozette2010lunar,neish2011surficial,campbell1997regolith,fa2013circular,Fa2018} \\

15 &
LRO Diviner Bolometric Temperature ($\rm T_{bol}$) &
LRO Diviner radiometer ($8-400\,\mu$m) &
15\,km/px; 90\textdegree S--90\textdegree N &
240\,m/px; 80--90\textdegree &
Diurnal surface-temperature sampling across 24 sub-solar-longitude states (polar maps split summer/winter) &
\cite{paige2010lunar,Williams2017,Williams2019} \\

16 &
LRO Diviner Ice Stability Depth &
Derived from LRO Diviner ($8-400\,\mu$m) + thermal modeling &
\textemdash{} &
240\,m/px; 80--90\textdegree &
Depth below the surface at which water ice is thermally stable to sublimation &
\cite{Schorghofer2020} \\

17 &
LRO Diviner Thermophysical Products &
Derived from LRO Diviner nighttime thermal modeling ($10-400\,\mu$m) &
240\,m/px; 70\textdegree S--70\textdegree N &
\textemdash{} &
Rock abundance (areal fraction of m-scale rocks); nighttime regolith $\rm T_{bol}$ anomaly; H-parameter (regolith density scale height / thermal inertia) &
\cite{powell2023high,hayne2017global} \\

18 &
Lunar Prospector Hydrogen Abundance &
LP Neutron Spectrometer &
\textemdash{} &
15\,km/px; 60--90\textdegree &
Near-surface hydrogen abundance to depths of tens of cm &
\cite{Lawrence2022,feldman1999lunar,feldman2001evidence} \\

19 &
GRAIL Free-Air Gravity &
Derived from GRAIL dual-spacecraft Ka-band tracking &
20\,km/px; 90\textdegree S--90\textdegree N &
20\,km/px; 60--90\textdegree &
Free-air gravity disturbance; basin and mascon structures, crustal heterogeneity; links surface to subsurface mass anomalies &
\cite{zuber2013gravity,konopliv2014high,lemoine2014grgm900c,Goossens2020,Park2025} \\

20 &
USGS Unified Geologic Map &
Apollo-era maps + LROC WAC morph.\ map + SLDEM2015 &
60\,m/px; 90\textdegree S--90\textdegree N &
60\,m/px; extends to poles &
43 unique geologic units identified based on surface texture, morphology, composition, rock type, and age &
\cite{Fortezzo2020} \\

\end{longtable}
}

\medskip
\noindent We describe each product below, noting where a polar-focused version differs from its global counterpart. Products released for only one region are described once.

\textit{LROC WAC Global Morphologic Mosaic}

The LROC WAC global morphologic mosaic provides a global, single-band (643 nm), medium-resolution (100 m/pix) basemap designed to emphasize surface morphology and texture (e.g., crater rim sharpness, wrinkle ridges, flow fronts, blocky ejecta, and subtle topographic shading). The product is assembled specifically under high-incidence illumination to enhance relief; typical incidence angles are reported as \textasciitilde{}55–75° with the average being 60°, which increases shadowing and makes morphologic boundaries easier to interpret than low-incidence albedo mosaics \cite{speyerer2011lunar,robinson2012exploring,wagner2015new}. The mosaic is released in tiles with equirectangular projection spanning mid-latitudes and polar stereographic caps poleward of 60°. Because the mosaic integrates observations acquired across multiple periods to manage illumination and coverage, the native per-image geometry (incidence, emission, and phase) is not a single fixed value. Instead, it varies by source frame and is tracked in the upstream WAC image metadata, while the mosaicing process reduces visible seams and balances brightness to create a consistent morphologic basemap \cite{speyerer2011lunar,robinson2012exploring,wagner2015new}.  In the polar regions, the WAC morphologic mosaics span $30\degree$ from both poles at 100 m/pix and are assembled from observations selected to emphasize topographic shading under the persistently grazing polar illumination \cite{wagner2015new,speyerer2013persistently}.

\textit{LROC WAC Normalized Reflectance Mosaics (321–689 nm plus 643 nm higher-resolution basemap)}

The LROC WAC normalized reflectance mosaics are seven-band global (or near-global) products that represent photometrically corrected reflectance (radiance factor, I/F) in the WAC color bands (321, 360, 415, 566, 604, 643, 689 nm) \cite{sato2017lunar}. These mosaics are explicitly intended for comparative spectral and compositional analysis by reducing brightness variations caused by changing illumination and viewing geometry. In each of the mosaics, each pixel is normalized to a standard geometry of incidence = 30°, emission = 0°, and phase = 30°, using a globally derived photometric function, similar to that of \cite{boyd2012lunar}, placing all observations on a common reference geometry and making band-to-band ratios and subtle spectral contrasts more robust. The empirically normalized 500 m/pixel resolution product spans 60°S–60°N and 0–360°E and was constructed from \textasciitilde{}137,400 WAC color images acquired over multi-year mapping, sampling a wide range of original incidence/emission/phase angles before normalization \cite{sato2017lunar,sato2014resolved}. The higher-resolution 643 nm product (100 m/pix) includes regions within $30\degree$ of the poles, providing a co-registered polar reflectance layer that reduces geometry-driven brightness variability relative to individual images and improves comparability across polar terrain units and integration with thermal and radar layers \cite{wagner2015new,speyerer2013persistently}.

\textit{Kaguya (SELENE) Multi-Band Imager (MI) Normalized Reflectance mosaics (414–1548 nm)}

The Kaguya (SELENE) Multi-Band Imager (MI) normalized reflectance mosaics provide global mapped reflectance in the VIS–NIR (visible through near-infrared) across bands centered near 415, 750, 900, 950, 1000, 1050, 1250, and 1550 nm \cite{lemelin2015lunar}, enabling analyses sensitive to lunar mafic mineral absorption behavior near a wavelength of \textasciitilde{}1 µm as well as broadband maturity and albedo trends. Instrumentally, MI acquires push-broom imagery with nominal spatial resolution of \textasciitilde{}20 m in the visible system and \textasciitilde{}62 m in the NIR system from the \textasciitilde{}100 km mapping orbit \cite{lemelin2015lunar}. Global mosaics are subsequently map-projected and commonly resampled to standardized grids for multi-layer analysis \cite{ohtake2008performance}, with a standard geometry of incidence = 30°, emission = 0°, and phase = 30°, and a resolution of 60 m/pixel. The distributed mosaics are topographically corrected MI reflectance, which is a critical distinction for lunar work because it reduces terrain-driven shading that can otherwise masquerade as compositional contrast in rugged regions \cite{lemelin2015lunar}. As with other global mosaics, the per-pixel original observing geometry varies across the contributing frames. However, the delivered MI mosaic products aim to suppress these effects via topographic/photometric correction so that spectral comparisons across latitude, terrain type, and illumination regimes are more physically interpretable \cite{lemelin2015lunar,lemelin2016global,lemelin2019compositions}.

\textit{Kaguya (SELENE) Multi-Band Imager (MI) Derived Mineralogy}

In addition to reflectance, Kaguya MI compositional and maturity layers, that invert the MI multi-band reflectance into physically interpretable properties, are derived \cite{lemelin2015lunar,lemelin2016global,lemelin2019compositions}. Five mineralogy maps including olivine, orthopyroxene (low-Ca pyroxene), clinopyroxene (high-Ca pyroxene), plagioclase, and FeO were created at a resolution of 60 m/pixel, along with plagioclase grain size and optical maturity (OMAT). Three specialized layers were derived to further analyze submicroscopic iron \citep{Trang2019}. These three 1 km/pixel maps focused on microphase Fe (mpFe), sub-microscopic Fe (smFe), and nanophase Fe (npFe). These maps are powerful because they move beyond qualitative spectral contrasts to provide standardized proxies for mineralogy and maturity at global scale, and they are produced from topographically corrected MI mosaics to reduce slope-induced albedo artifacts \citep{Trang2019}.

\textit{Kaguya (SELENE) Spectral Profiler (SP) Derived Mineralogy}

To provide polar compositional constraints beyond broadband imaging, Kaguya (SELENE) Spectral Profiler (SP) derived mineral and maturity maps are used. SP is a nadir-looking VIS–NIR spectrometer spanning roughly 0.5–2.6 µm with \textasciitilde{}6–8 nm spectral sampling, designed to resolve diagnostic mineral absorption structure at lunar surface compositions \cite{haruyama2008global,yamamoto2014calibration}. These 1 km/pixel mineral maps consist of olivine,  orthopyroxene (low-Ca pyroxene), clinopyroxene (high-Ca pyroxene), plagioclase, and FeO, as well as nanophase Fe (npFe) and optical maturity (OMAT).

\textit{LROC WAC Titanium (TiO$_2$)}

The LROC WAC TiO$_2$ map provides a global estimate of titanium dioxide abundance (wt\%) derived from the WAC UV/VIS ratio, specifically using the 321/415 nm band ratio relationship calibrated against lunar sample constraints and cross-compared to legacy TiO$_2$ products \cite{sato2017lunar}. The TiO$_2$ derived map has a near-global coverage of 70°N–70°S, 0–360°E and resolution of 400 m/pixel \cite{sato2017lunar}. This dataset is intended to support mare basalt compositional discrimination and regional stratigraphic interpretations. However, it should be treated as a model-derived abundance layer whose uncertainties can increase in optically immature materials (e.g., fresh rays and very young ejecta) where space-weathering state perturbs UV/VIS behavior \cite{sato2017lunar}.

\textit{LOLA/Kaguya TC Merged Topography and Derivatives (Slope, Aspect, Aspect Sine/Cosine)}

To represent lunar shape and local terrain geometry, we use the merged LOLA–Kaguya Terrain Camera (TC) DEM product (commonly referenced as SLDEM2015), which co-registers Kaguya TC stereo topography to the LOLA geodetic framework to produce a consistent elevation model across the globe where TC coverage is available \cite{barker2016new}. This DTM derived map has a near global coverage of 60°N–60°S, 0–360°E and resolution of 60 m/pixel with a typical vertical accuracy reported as \textasciitilde{}3–4 m \cite{barker2016new}. We compute standard geometric derivatives: slope (local gradient magnitude) and aspect (downslope azimuth), along with aspect sine and aspect cosine layers that encode aspect direction in a manner that avoids the 0°/360° discontinuity, all at 60 m/pixel resolution \cite{barker2016new}. Because these are purely topographic descriptors, they do not carry incidence/emission/phase metadata. Their role is instead to (i) quantify morphometry directly and (ii) support illumination-aware interpretation and modeling of optical and thermal datasets that are strongly slope and aspect dependent \cite{barker2016new}. The original pre-mosaicked data consist of 43,200 DTMs from TC and 4.5 billion altimetric surface heights from LOLA \cite{barker2016new}. In the polar regions, these geometric layers are foundational for interpreting illumination and temperature fields because small changes in local horizon, slope, and aspect strongly control solar visibility and thermal environments near the poles \citep{barker2023,barker2025large}.

\textit{LOLA Roughness}

The LOLA roughness map is a measure of background surface texture at a 50 m baseline that complements DTM-derived slope/aspect and provides an independent constraint on terrain physical state (e.g., blockiness, regolith development, and degradation) \cite{kreslavsky2013lunar,neumann2015}. The roughness static map is constructed directly from LOLA elevation footprints rather than from an interpolated gridded DTM, leveraging the high internal precision of the laser altimetry to map texture in a way that is robust to inter-track spacing and orbit-to-orbit absolute offsets \cite{kreslavsky2013lunar,neumann2015}. 
For this product, roughness is estimated from the standard deviation of height residuals of the five adjacent spots returned from a single LOLA laser pulse after removing a plane surface representing the local slope \cite{neumann2015}. 
The resulting global 1 km/pixel map provides interpretable geologic context and age distinctions for the lunar surface, and the same 1 km/pixel product is used in the polar regions \cite{kreslavsky2013lunar,neumann2015}.

\textit{LOLA Average Sun Illumination and Permanently Shadowed Regions (PSRs)}

Two LOLA-derived polar products explicitly quantify the polar illumination environment: permanently shadowed regions (PSRs) at 20 m/pixel \cite{barker2023,barker2025large} and the average Sun illumination at 120 m/pixel \cite{mazarico2011illumination}. These layers are generated by numerical illumination modeling using LOLA topography over a full lunar precession cycle (\textasciitilde{}18.6 years) to compute long-term solar visibility statistics and identify terrain that remains persistently unilluminated \cite{mazarico2011illumination}.

\textit{LOLA 1064 nm Normal Albedo}

The LOLA 1064 nm normal albedo layer provides a polar-capable reflectance proxy that is insensitive to the limitations of Sun-illuminated optical imaging at extreme latitudes. LOLA measures the backscattered energy from its 1064 nm laser pulse which allows it to map reflectivity in a geometry that is effectively zero phase. This yields a consistent albedo characterization that is independent of illumination and observation geometry even at high latitudes where conventional photometry is challenging \cite{Lemelin2016}. These 1 km/pixel maps are particularly relevant for studies linking polar brightness to thermophysical thresholds and PSR environments.

\textit{LRO Mini-RF Radar Reflectivity and Circular Polarization Ratio (CPR)}

To characterize near-surface physical properties that are not uniquely constrained by optical reflectance, we use LRO Mini-RF radar mosaics, including radar reflectivity (backscatter) and the circular polarization ratio (CPR) \citep{fassett2024}. Mini-RF is a hybrid-polarimetric synthetic aperture radar (SAR) instrument that can acquire data at S-band (\textasciitilde{}12.6 cm) and X/C-band (\textasciitilde{}4.2 cm) \cite{raney2007hybrid,nozette2010lunar,raney2010lunar}. Radar reflectivity provides sensitivity to wavelength-scale roughness and dielectric contrasts, while CPR highlights scattering behavior linked to blockiness, multiple scattering, subsurface structure, and composition \cite{nozette2010lunar,neish2011surficial,campbell1997regolith}. As with other strip-built global mosaics, local incidence and look geometry vary by acquisition and are tracked in the underlying observation metadata, while the mosaic products correct for these variations and are geodetically-controlled, providing map-projected layers suitable for global comparative analysis at 90 m/pixel resolution \citep{fassett2024}.  The polar reflectivity and CPR mosaics (90 m/pixel) are particularly valuable in and around PSRs, where optical reflectance is absent, extremely weak, or strongly geometry-limited. Both are geodetically-controlled and corrected for topography and look geometry, and differ from the global mosaics mainly in projection \cite{nozette2010lunar,spudis2013evidence}. Importantly, CPR anomalies are not uniquely diagnostic of water ice. Elevated CPR can also arise from rough, blocky, or volume-scattering surfaces unrelated to volatiles \cite{fa2013circular,nozette2010lunar,Fa2018}. For that reason, CPR is treated as a physical-property/classification layer whose primary role is contextualization (e.g., separating rough/blocky terrains from smoother regolith units), analyzed jointly with temperature and illumination constraints rather than as a standalone ice-detection map \cite{fa2013circular,nozette2010lunar}.

\textit{LRO Diviner Bolometric Temperature and Ice Stability Depth}

The Diviner bolometric temperature $\rm T_{bol}$ maps provide gridded regolith temperature estimates derived from Diviner thermal channels 3–9 \citep{paige2010lunar}. This family of maps is distributed as global cylindrical maps gridded at $0.5\degree$/pixel ($\sim15$ km/pixel) \citep{Williams2017}. In this dataset suite, the sequence of bolometric temperature layers labeled 0, 15, ..., 345 is treated as a standardized sampling of diurnal forcing at sub-solar longitude increments of $15\degree$. To create each map, the diurnal temperature curve from Diviner observations within each 0.5\degree{} pixel was interpolated \citep{Williams2017}. Some artifacts exist due to gaps in spatial or temporal coverage. These are typically most apparent near the dawn and dusk terminators or at low-to-mid latitudes. Bolometric temperature measures the spectrally integrated flux of infrared radiation emitted by the surface \citep{paige2010lunar}, and it is important for quantifying the overall heat balance of the surface, which depends on the regolith's thermophysical properties \citep{Williams2017}. The polar $\rm T_{bol}$ maps use the same sub-solar-longitude binning but at higher resolution (240 m/pixel) and are subdivided into summer and winter seasons to account for the grazing polar illumination angles and the Moon's non-zero obliquity \citep{Williams2019}. In addition, the Diviner ice stability depth map integrates Diviner temperature constraints with thermal/volatile stability modeling to estimate where water ice is thermally stable at and below the surface under polar conditions at 240 m/pixel \cite{Schorghofer2020}.

\textit{LRO Diviner Thermophysical Products (Rock Abundance, Nighttime Regolith Temperature Anomaly, H-Parameter)}

To capture temperature and thermophysical properties of the upper regolith, Diviner products translate multi-channel thermal IR radiance into physically interpretable maps are used. The rock abundance map estimates the areal fraction of meter-scale rocks, and the nighttime regolith temperature map highlights departures from slope-adjusted midnight temperatures typical for a given latitude \cite{powell2023high}. Both maps were derived from nighttime Diviner channels 6–9 and made by fitting nighttime thermal radiance with a two-component mixture model (rocks and fine regolith), exploiting the higher thermal inertia of rocks that remain anomalously warm at night \cite{powell2023high}. The H-parameter is a regolith structural parameter that governs how density and thermal conductivity increase with depth \cite{hayne2017global}, which serves as a proxy for thermal inertia. The derived global map was constructed by fitting Diviner temperature observations with a thermal model that links subsurface density profiles to observed nighttime temperatures, enabling interpretations in terms of regolith packing state and impact-driven evolution \cite{hayne2017global}. All three thermophysical static maps are 240 m/pixel and span 70°N–70°S.

\textit{Lunar Prospector Hydrogen Abundance}

The Lunar Prospector Neutron Spectrometer–derived hydrogen abundance map (15 km/pixel) measures epithermal neutron suppression associated with near-surface hydrogen (typically sampling depths of tens of centimeters) \citep{Lawrence2022}. The polar hydrogen enhancements detected by Lunar Prospector are a cornerstone observation that captures volatile-related geochemical context and motivates cold-trap and volatile-retention hypotheses \cite{feldman1999lunar,feldman2001evidence}. This dataset provides a compositionally-grounded complement to Diviner thermal and LOLA illumination layers that is independent of solar illumination and thus probes PSRs.

\textit{GRAIL Free-Air Gravity Disturbance}

The GRAIL free-air gravity disturbance 20 km/pixel static map represents the Moon’s gravity field at spatial scales that resolve basin and mascon structures, and other heterogeneities of the lunar interior structure \cite{zuber2013gravity}. Analysis of GRAIL’s dual-spacecraft Ka-band inter-satellite tracking enabled gravity field mapping at substantially higher resolution than pre-GRAIL datasets, resolving many surface and subsurface features on the Moon (particularly on the far side) and it continues to provide a reference global geophysical context layer complementary to morphology and composition \cite{zuber2013gravity,konopliv2014high, lemoine2014grgm900c,Goossens2020, Park2025}. In this workflow, the inclusion of a gravity disturbance map provides a link between surface expressions (e.g., volcanic provinces, basin rings, terrains) observed in the optical/radar/thermal datasets and coherent subsurface mass distributions observed in the gravity field. The polar workflow uses the polar regions of this same global product.

\textit{USGS Unified Geologic Map of the Moon}

The 2020 USGS Unified Geologic Map of the Moon \citep{Fortezzo2020} was used as input data representing differentiated geologic units. This map represents a global mapping of discrete geologic units across the entire lunar surface at a scale of 1:5,000,000. The map consists of 43 unique geologic units identified based on surface texture, morphology, composition, rock type, and age. The map includes units which span all lunar geologic time periods including the Pre-Nectarian, Nectarian, Imbrian, Eratosthenian, and Copernican periods. The units range greatly in size from the largest, which encompass surface mare deposits to small units surrounding well-preserved impact structures. The Unified Geologic Map of the Moon combines Apollo-era regional maps with modern data and surface analysis to present a consistent global summary of lunar surface geology and is the standard global geologic map used by the lunar science community \citep{Fortezzo2020}.

Together, the global and polar static maps provide a consistent, multi-modal representation of lunar geology, geophysics, and geochemistry, including composition-sensitive proxies, terrain geometry, thermophysical state, and illumination environment. \dataset uses these layers, together with the selected NAC and WAC imagery described above, as the raw material for the pre-training corpus and the benchmark tasks. We next describe the preprocessing that converts the raw image products into calibrated, map-projected rasters and harmonizes the static maps onto common grids, followed by the construction of the ML-ready pre-training dataset (Section~\ref{sec:construction}) and the application benchmark datasets built on top of it (Section~\ref{sec:benchmarks}).

\subsection{Lunar Data Preprocessing}
\label{subsec:preprocessing}

The preprocessing of Narrow Angle Camera (NAC) and Wide Angle Camera (WAC) images captured by the Lunar Reconnaissance Orbiter Camera follows a structured pipeline designed to convert the raw Experiment Data Record (EDR) mission image products, stored in the PDS archive, into radiometrically corrected and map-projected, or orthorectified, images used to create a dataset suitable for machine learning. ISIS, short for Integrated Software for Imagers and Spectrometers\footnote{https://doi.org/10.5066/P13YBMZA}, is a planetary image processing software package used to convert raw images into data suitable for scientific analysis. The \texttt{lronac2isis} and \texttt{lrowac2isis} modules convert the original \texttt{.IMG} files into the ISIS \texttt{.cub} format. This step makes the raw images compatible with the image processing tools available in the ISIS software package while preserving the instrument metadata. After successful conversion of the raw files into the ISIS \texttt{.cub} format, the \texttt{spiceinit} module is used to attach spatial geometry, camera position, and other geometric information to the metadata, allowing each image to be spatially interpreted as an  observation of the lunar surface.

Once the geometric context has been established, the NAC and WAC images undergo instrument-specific radiometric corrections specific to each instrument. The ISIS  \texttt{lronaccal} and \texttt{lrowaccal} commands are used to radiometrically calibrate the raw sensor response, removing or compensating for artifacts or errors caused by the instruments so that the image values accurately represent the observed surface signal. The \texttt{lronacecho} and \texttt{lrowacecho} commands are then used to reduce a known and well quantified echo effect that can cause repeated signals between detector columns.

After applying all corrections to the NAC and WAC images and converting them into processed ISIS \texttt{.cub} files, the Ames Stereo Pipeline\footnote{https://zenodo.org/records/15298734} \citep[ASP]{beyer2018ames} \texttt{mapproject} utility projects and orthorectifies the images using a lunar digital elevation model, in this case the LOLA-Kaguya DEM. This corrects the final rasters for both sensor geometry and distortions caused by terrain. The coordinate reference system (CRS, projection) of the image is selected based on the Lunar Transverse Mercator (LTM) zone in which it falls \citep[][see Sec. \ref{ia:tiling}]{McClernan2025}. Following map projection, \texttt{gdal\_translate} converts the rasters into Cloud Optimized GeoTIFFs, which are then used to create the ML ready dataset.

The global static maps are first harmonized to an equidistant cylindrical projection (IAU\_2015: 30110), after which 90 virtual rasters (VRTs) are created to reproject the data into the corresponding LTM zones. The polar static maps are initially provided in polar stereographic projections (IAU\_2015:30130 and 30135) and are reprojected using VRTs into the Lunar Polar Stereographic (LPS) system \citep{McClernan2025}.

\subsection{Pre-training Dataset Construction}
\label{sec:construction}
\label{sec:image_anchored_pipeline}

The \dataset pre-training dataset is designed for large scale multimodal self-supervised learning. It is constructed using an \emph{image-anchored pipeline} that uses a common upstream substrate: (i) the 90 LTM and 2 LPS VRTs prepared in Section~\ref{subsec:preprocessing}, (ii) the LROC EDR geodatabase, and (iii) a single registry of per channel value ranges.

The image-anchored pipeline defines tiles inside individual
LROC EDR canvases. A single sliding window generates many
overlapping patches of 512 x 512 pixels per EDR, with all co-registered
modalities snapped to a shared per-tile bounding box. This structure is important for ML because each training sample contains co-registered modalities snapped to the same exact bounding box, allowing the model to learn meaningful cross-modal relationships from spatially corresponding observations. Anchoring tiles to NAC or WAC image canvases also preserves the optical observation context, and overlapping EDR coverage can expose the model to the same or nearby terrain under different illumination conditions. The pipeline is
implemented as two parallel tracks, \emph{low-resolution} (WAC-anchored) and \emph{high-resolution} (NAC-anchored), that follow a common two-pass construction pattern but differ in their optical anchor, tile resolution, and modality requirements. Both tracks reuse the perstrip Lunar Transverse Mercator (LTM) and per-pole Lunar Polar Stereographic (LPS) VRTs prepared in Section~\ref{subsec:preprocessing} and the LROC EDR geodatabase, and apply a per-channel value range clipping convention. For each modality, physical ranges for every band are defined in a per-modality registry and applied at write time. NaN values are preserved as true data gaps, and categorical layers are exempt. The public configuration parameters and their default values, as well as the required modalities for each track, are summarized in Table~\ref{tab:track-comparison}.

\subsubsection{Tiling Strategy: EDR-Anchored Sliding Window}
\label{ia:tiling}

Each LROC EDR is read directly through its strip-LTM or
per-pole LPS projected source. VIS+UV WAC products and NAC
products are pre-warped per strip upstream, so the EDR's pixel
canvas is already in the target tile CRS. Tile geometry is then defined by a
sliding window over that canvas with two configurable parameters,
the per-tile pixel side length (i.e. tile size dimension) and the rejection stride
(Table~\ref{tab:track-comparison}).

\paragraph{Adaptive sliding window.}
The window enumerates positions in raster order and applies
an adaptive advance policy: when a candidate window contains no
source NaN and yields all required modalities, the column index is
advanced by the full tile size, which produces disjoint tiles in clean data. When the window is rejected, most commonly because the source canvas
contains NaN over the window footprint, or, in the low-resolution
track, because a required co-registered modality returned more than
$90\%$ NaN, the column index is advanced by the rejection stride. This stride is typically a fraction of the tile size so that the next
valid position is recovered quickly. The same policy applies to row
advancement: rows in which at least one tile was accepted advance by
the tile size, otherwise by the stride. NAC EDRs frequently begin
with a header strip of NaN rows that a naive scan would otherwise traverse one pixel at a time. The high-resolution track prefaces the main scan
with a stride-only initial scan that locates the first valid window
before switching to the adaptive policy. Because it is difficult to find tiles without missing data at the poles, the pipeline performs a second pass for polar images, extracting tiles with any valid data without applying a NAN fraction threshold, or expanding iteratively from existing polar tiles in all four directions rather than scanning the full raster. This ensures complete coverage while avoiding traversal of empty regions.

\paragraph{LTM-strip identification.}
Each emitted tile is tagged with the LTM-zone or LPS-cap
code containing it. The tile's projected bounding box is densified and unprojected to lon/lat, and each corner of the unprojected bounding box is mapped to its LTM/LPS
code. When the upper left and lower right corners disagree, indicating the tile overlaps multiple zones, the tile is dropped. This ensures clear boundaries for splitting the dataset into train, val, and test sets by LTM-/LPS-zone without any spatial overlap across datasets, which is important to confirm that there is no data leakage between training and evaluation splits.

\subsubsection{CRS-Handling Protocol}
\label{ia:crs}

Because the EDR canvas is already in the strip-LTM or
per-pole LPS target CRS, tile extraction never reprojects the EDR
itself. Instead, WAC VIS and NAC tile data are simply sliced out of
the source raster at the window's pixel coordinates. The
CRS handling work is therefore confined to attaching co-registered
static layers and, in the low-resolution track, the per image WAC
UV) at the tile's exact bounds and a snapped resolution.
\begin{enumerate}[nosep,leftmargin=1.4em]
  \item \textbf{Snapped per-layer resolution.} For each layer with
        native resolution $r_\ell$, the per-tile pixel count is
        $N_\ell = \mathrm{round}(\mathrm{extent}/r_\ell)$ and the
        snapped resolution is $\mathrm{extent}/N_\ell$. Snapping
        guarantees an integer pixel count along the tile and exact
        alignment with the reference WAC VIS or NAC grid, so all
        modalities of a given tile share corner pixels even though
        their native resolutions span four orders of magnitude. This results in slight resolution modifications to the paired layers, as outlined in Table~\ref{tab:modality-coverage-and-resolution}.
  \item \textbf{Padded clip-then-reproject.} Each layer is read
        from its per-strip LTM or LPS VRT, keyed by the tile's
        \texttt{LTM\_CODE}. Source bounds are derived by
        transforming tile bounds back to the source CRS with edge
        densification (as described in Rule~3 below) and padding by one source
        pixel on every side. The padded clip is then reprojected
        onto the destination grid at the snapped resolution.
        Bilinear is the default resampling. Categorical layers, such as the unified geologic map, request nearest neighbor
        resampling through the layer's modality registry entry.
  \item \textbf{Four-corner bounding-box arithmetic.} All bounds transforms between
        CRSs densify each edge, with $21$ points per edge by default, to
        avoid minimum or maximum value swaps under nonlinear projections.
\end{enumerate}
A per-layer NaN gate caps the fraction of NaN pixels admitted: if
more than $90\%$ of the extracted clip is NaN, the layer is
recorded as missing for the tile and the actual NaN fraction is
written to the catalog. VRT handles are cached for each process so that each strip VRT is opened only once per worker.

\paragraph{Static-map redundancy across per-image tiles.}
Each co-registered static map layer is clipped anew for
every accepted EDR tile in this view, so a ground patch covered
by $m$ admitted EDRs is materialized $m$ times across the
per-modality netCDF files. This choice trades static-map storage for
alignment to the per-EDR pixel grid.

\subsubsection{Two-Pass EDR-Anchored Tile Construction}
\label{ia:two-pass-construction}

Both tracks follow a two-pass construction pattern. In Pass 1, an optical EDR product (WAC VIS for low-resolution, NAC for high-resolution) defines candidate tile geometry through the adaptive sliding-window procedure described in Section~\ref{ia:tiling}. In Pass 2, co-registered modalities are attached at the accepted tile bounds using the CRS-handling protocol described in Section~\ref{ia:crs}. The tracks differ primarily in their optical anchor, Pass 1 requirements, and the modalities attached in Pass 2 (Table~\ref{tab:track-comparison}).

\paragraph{Pass 1: Optical Anchor Extraction}
\label{ia:pass1-optical}
\subparagraph{Common framework.}
Both tracks begin with a JSON manifest of pre-filtered EDR products, processed in chunks or array-job slices. 
Per-product viewing geometry metadata is joined from the LROC EDR geodatabase, including \texttt{INCIDENCE\_ANGLE}, \texttt{EMISSION\_ANGLE}, \texttt{PHASE\_ANGLE}, \texttt{SUB\_SOLAR\_GROUND\_AZIMUTH},
 \texttt{SUB\_SOLAR\_LATITUDE}, \texttt{SUB\_SOLAR\_LONGITUDE}, \texttt{CENTER\_LATITUDE}, \texttt{CENTER\_LONGITUDE}, and the product's nominal \texttt{RESOLUTION}. 

The worker opens each EDR raster lazily and applies the adaptive sliding window of Section~\ref{ia:tiling} over the optical canvas. For each candidate window:
\begin{itemize}[nosep,leftmargin=1.2em]
  \item The optical slice is extracted at the window's pixel coordinates.
  \item Non-polar windows containing any NaN are rejected, while polar windows are accepted if they contain any valid data to maximize the number of training samples in these regions.
  \item The LTM-zone or LPS-cap code is resolved from the tile's corners, and tiles that span multiple zones are rejected.
  \item If all Pass 1 requirements are met, the tile is saved and a row containing the geometry, bounds, and viewing geometry fields is appended to the per-tile catalog.
\end{itemize}

Records are flushed in batches into part-Parquet files. A bootstrap mode scans existing output Parquet files for already-processed product IDs and skips those products when processing restarts.

\subparagraph{High-resolution track specifics.}
\label{ia:pass1-highres}
The high-resolution track is anchored to NAC EDRs. The NAC EDR list is provided through a JSON manifest that is divided across array jobs. The worker skips NACs whose usable dimensions (in pixels) are smaller than the configured tile size, locates the first valid window via an initial stride scan, and then runs the adaptive sliding window of Section~\ref{ia:tiling}. For each candidate:
\begin{itemize}[nosep,leftmargin=1.2em]
  \item Non-polar windows containing any NaN are skipped, while polar windows containing only NaN are skipped.
  \item The LTM-zone code is resolved from the tile's corners, and the tile is skipped if it crosses multiple LTM zones.
  \item Otherwise, the NAC slice is saved as a single-band netCDF in the \texttt{nac/} sub-directory of the output root.
  \item One row is appended to the per-tile catalog containing the \texttt{NAC\_TILE} path, primary \texttt{LTM\_CODE}, corner lon/lat, projected bounds, and the viewing geometry fields from the EDR geodatabase.
\end{itemize}

Modality attachment is deferred to Pass 2 so that the geometry pass remains focused on the most I/O-bound step, reading each large NAC rasters once, while allowing modalities to be processed independently in parallel. The NAC images used for the high-resolution dataset are drawn from the NAC PHO sites and the NAC orthorectified images used to create the LRO 3 m DTM. As a result, these NAC images are co-registered with one another, their associated DTMs, and other DTM-derived modalities.

\begin{figure}[!htbp]
    \centering
    \includegraphics[width=\textwidth]{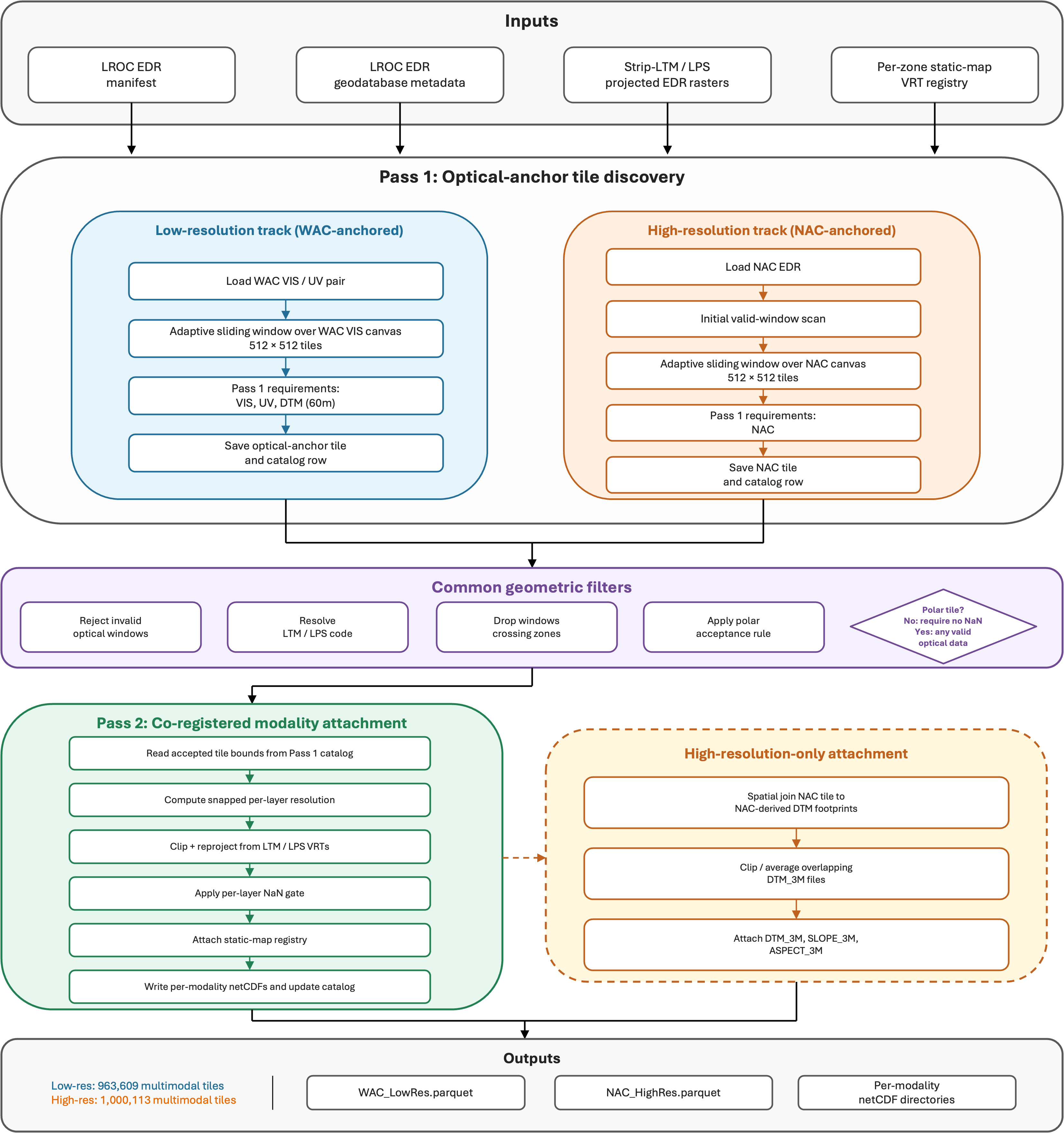}
    \caption{Tile extraction pipeline for the image-anchored low-resolution and high-resolution tracks.}
    \label{fig:image-anchored-pipeline}
\end{figure}

\paragraph{Pass 2: Modality Attachment.}
\label{ia:pass2-modalities}

Both tracks begin Pass 2 after Pass 1 is complete and attach co-registered modalities at the bounds of each optical anchor tile using the CRS-handling protocol described in Section~\ref{ia:crs}. For each tile:
\begin{itemize}[nosep,leftmargin=1.2em]
  \item The tile is opened to obtain its bounds, the LTM/LPS zone is read from the Pass 1 catalog, and the snapped resolution is calculated.
  \item Each modality is extracted at the tile's bounds and snapped resolution from per-zone LTM/LPS VRTs.
  \item The per-layer NaN gate rejects layers with more than $90\%$ NaN. Rejected layers are recorded in the catalog as missing.
  \item Single-band layers are extracted directly, while multi-band layers (Kaguya MI mineralogy, Kaguya MI multispectral, WAC normalized reflectance, Kaguya SP polar mineralogy, MI space-weathering Fe) are stacked along a band dimension.
  \item For multi-phase Diviner layers (\texttt{TBOL}, \texttt{TBOL\_POLES}), all available phases are loaded as separate bands. \texttt{TBOL} contains all 24 sub-solar longitude snapshots (15$\degree$ steps), while \texttt{TBOL\_POLES} contains all 24 sub-solar longitude ranges for both summer and winter, resulting in 48 total bands. The product's sub-solar longitude is binned to identify the closest phase, which is duplicated as an additional "closest" band for convenience, yielding 25 bands for non-polar tiles and 49 bands for polar tiles.
  \item If a multi-band layer is missing any band, the entire layer is recorded in the catalog as missing for the tile. 
  \item Outputs are written as netCDF files within per-modality subdirectories, and the catalog is updated with the corresponding \texttt{\{MODALITY\}\_TILE} paths and \texttt{\{MODALITY\}\_FRACTION\_NULL} values.
\end{itemize}

Both tracks attach the same static-map registry (e.g., \texttt{DTM/DTM\_60M}, \texttt{GEOMAP}, \texttt{TREG}, \texttt{ROUGHNESS}, \texttt{MINIRF\_CPR}, \texttt{TIO2}, \texttt{WAC\_MOSAIC}, \texttt{MI\_MINERALOGY}, \texttt{TBOL}, \texttt{TBOL\_POLES}, \texttt{GRAVITY}, etc.). Polar-only modalities include \texttt{PSR}, \texttt{AVG\_ILLUM}, \texttt{ALBEDO}, \texttt{DICE}, and \texttt{HYDROGEN}. Non-polar only modalities include \texttt{TIO2}, \texttt{ROCK\_ABUNDANCE}, and \texttt{HPAR}. The final count of netCDF tiles per modality is provided in Table~\ref{tab:modality-coverage-and-resolution}.

\subparagraph{High-resolution track: additional NAC-derived DTM attachment.}
\label{ia:pass2-highres}
In addition to the static-map registry, the high-resolution track also attaches NAC-derived terrain products (\texttt{DTM\_3M}, \texttt{SLOPE\_3M}, \texttt{ASPECT\_3M}). A preprocessing script performs a spatial join to associate NAC-derived DTMs at 3 m resolution with each NAC tile row in the Parquet. Each NAC tile's lon/lat corner coordinates are converted to a geographic bounding box, and the native CRS bounds of all DTM files are transformed into the lunar geographic CRS (IAU\_2015:30100). A GeoPandas spatial join using the \texttt{intersects} predicate then identifies the DTM footprints that overlap each NAC tile. Results are grouped by tile and the matching DTM file paths are aggregated into a list in the \texttt{dtm\_file} column, yielding zero, one, or multiple DTM files per NAC tile depending on spatial coverage. The DTM TIFF(s) associated with each NAC tile are clipped to the tile bounds and averaged using a NaN-aware mean where multiple DTMs overlap. Slope and aspect are read from the corresponding derived-product TIFFs for each DTM. Figure~\ref{fig:image-anchored-pipeline} outlines the workflow to extract the tile modality bundles.

\paragraph{Example Outputs}
\label{ia:examples}

\begin{figure}[!htbp]
\caption{Example of low-resolution non-polar (\textit{top}) and polar (\textit{bottom}) samples, displaying all modalities side by side. Plots include the first channel of each modality at 51.2 x 51.2 km resolution, or 512 x 512 WAC visible pixels.
Certain modalities, such as \texttt{PSR} and \texttt{ALBEDO}, only exist at the poles and are thus missing from the non-polar sample. Similarly, other modalities, such as \texttt{TIO2} and \texttt{ROCK\_ABUNDANCE}, do not exist at the poles.}
\label{fig:low_res_modalities}
\centering
\begin{subfigure}{\textwidth}
    \centering
    \caption{Non-Polar Sample Modalities}
    \label{fig:nonpolar_low_res}
    \includegraphics[width=0.8\textwidth]{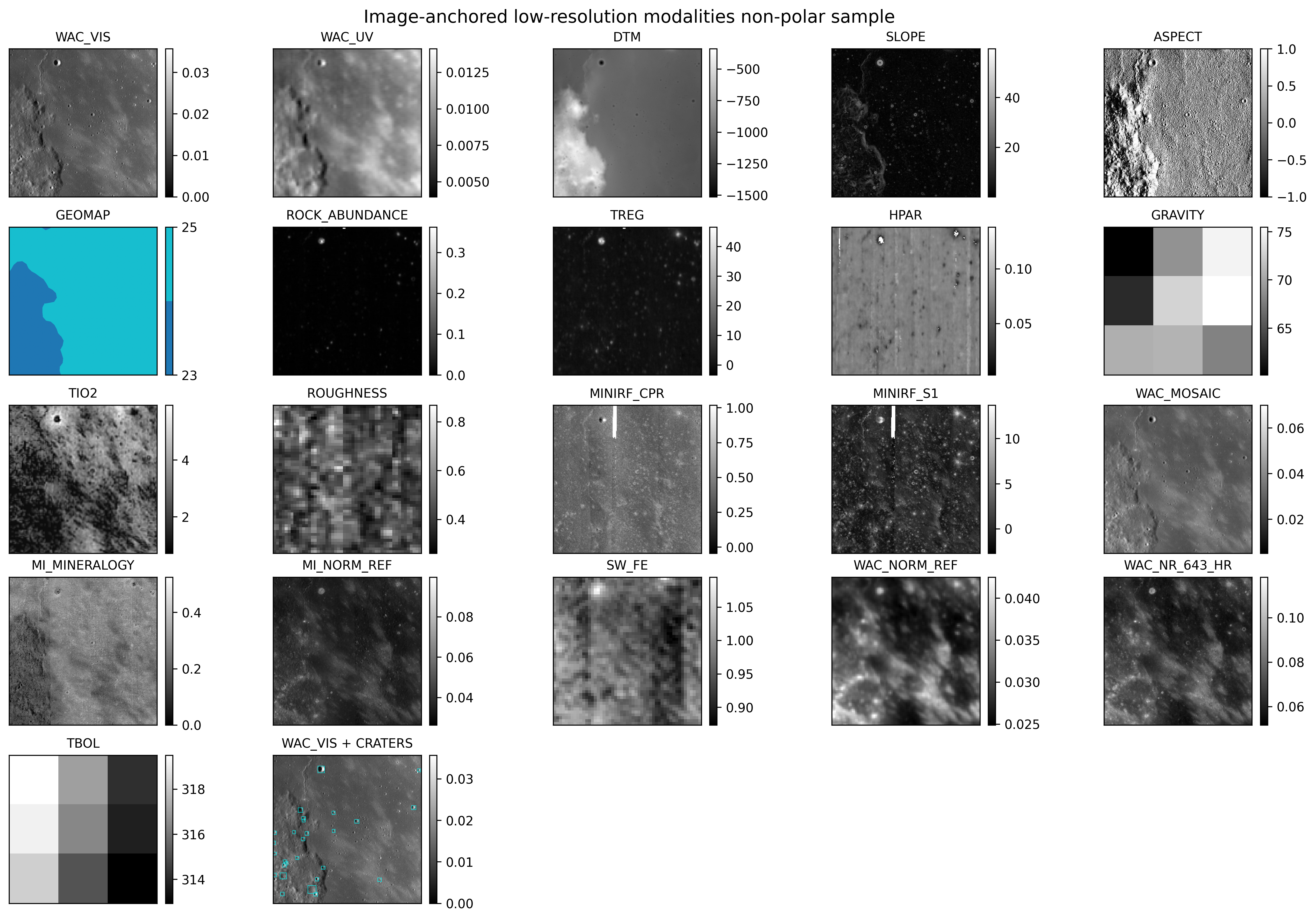}
\end{subfigure}

\vspace{1em}

\begin{subfigure}{\textwidth}
    \centering
    \caption{Polar Sample Modalities}
    \label{fig:polar_low_res}
    \includegraphics[width=0.8\textwidth]{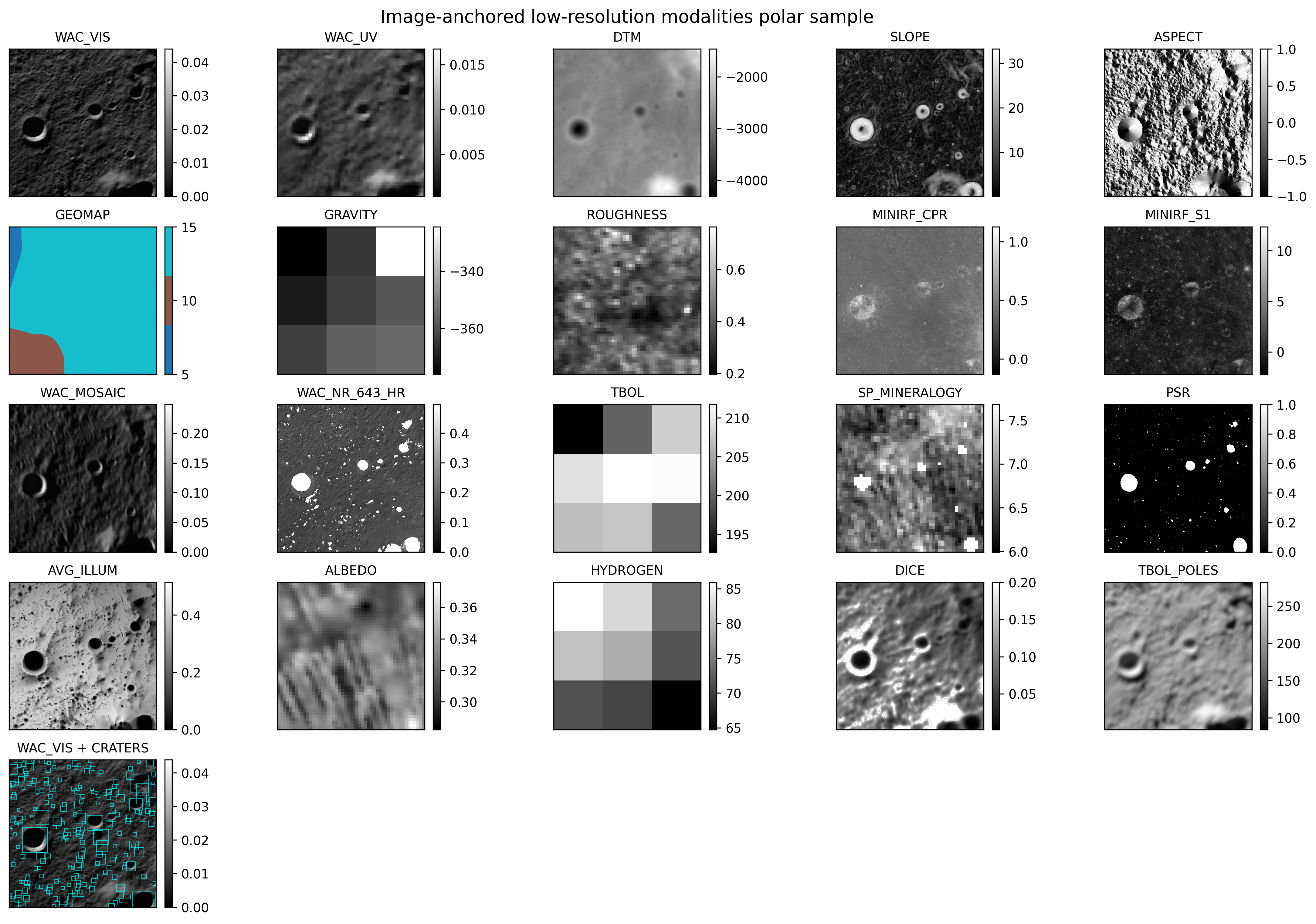}
\end{subfigure}
\end{figure}

Figure~\ref{fig:low_res_modalities} shows an example of a low-resolution non-polar sample and a polar sample, displaying all modalities side by side. Plots include the first channel of each modality at 51.2 km $\times$ 51.2 km resolution, corresponding to 512 $\times$ 512 WAC visible pixels. Certain modalities, such as \texttt{PSR} and \texttt{ALBEDO}, only exist at the poles and are therefore missing from the non-polar sample. Similarly, other modalities, such as \texttt{TIO2} and \texttt{ROCK\_ABUNDANCE}, do not exist at the poles.

\begin{figure}[!htbp]
\caption{Example of high-resolution non-polar (\textit{top}) and polar (\textit{bottom}) samples, displaying all modalities side by side. Plots include the first channel of each modality tile at 512 x 512 m resolution, or 512 x 512 NAC pixels.
Certain modalities, such as \texttt{PSR} and \texttt{ALBEDO} only exist at the poles and are therefore missing from the non-polar sample. Similarly, other modalities, such as \texttt{TIO2} and \texttt{ROCK\_ABUNDANCE} do not exist at the poles.}
\label{fig:high_res_modalities}
\centering
\begin{subfigure}{\textwidth}
    \centering
    \caption{Non-Polar Sample Modalities}
    \label{fig:nonpolar_high_res}
    \includegraphics[width=0.8\textwidth]{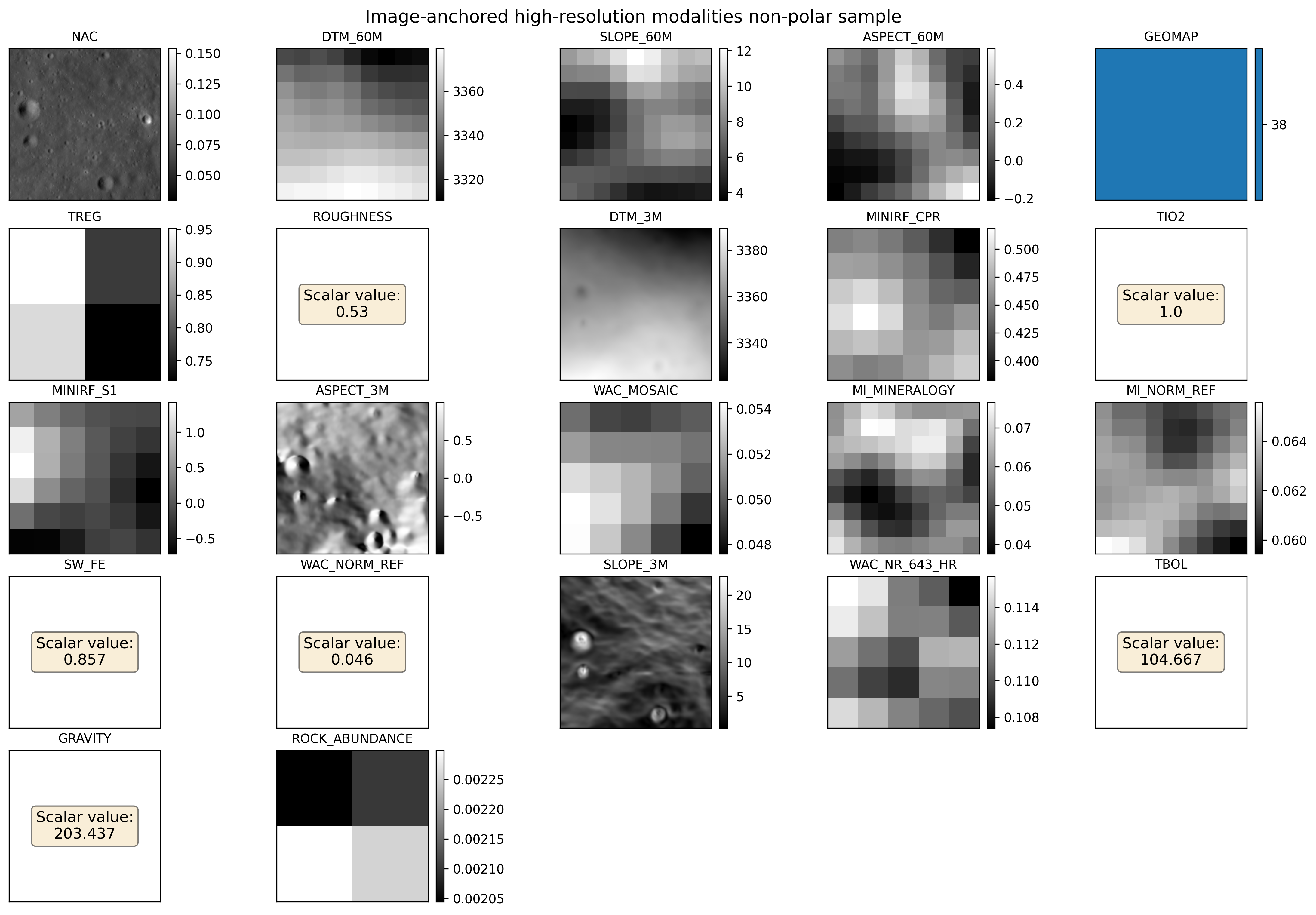}
\end{subfigure}

\vspace{1em}

\begin{subfigure}{\textwidth}
    \centering
    \caption{Polar Sample Modalities}
    \label{fig:polar_high_res}
    \includegraphics[width=0.8\textwidth]{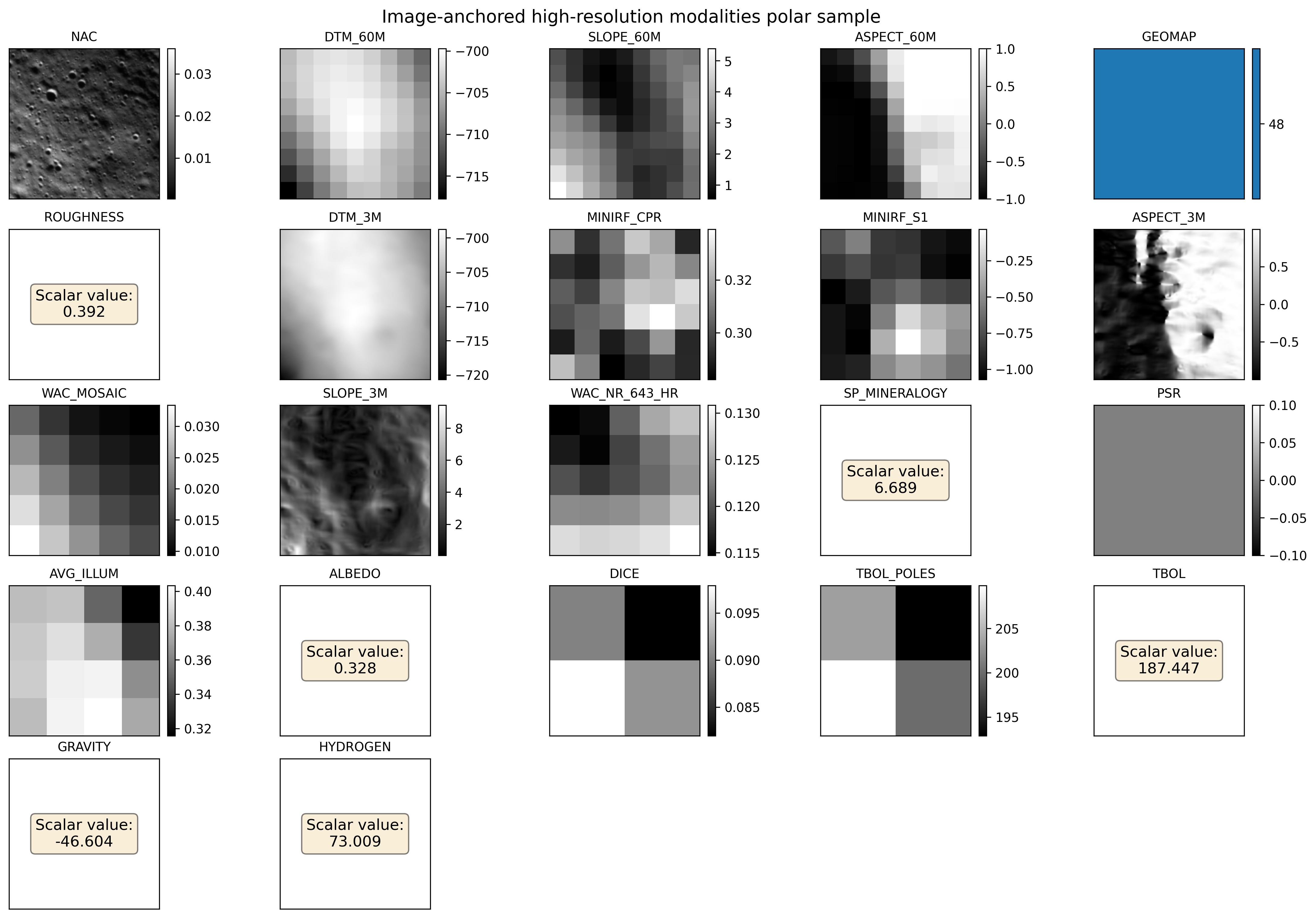}
\end{subfigure}
\end{figure}

Figure~\ref{fig:high_res_modalities} shows an example of a high-resolution non-polar and polar sample, displaying all modalities side by side. Plots include the first channel of each modality tile at 512 m $\times$ 512 m resolution, corresponding to 512 $\times$ 512 NAC pixels. The NAC tiles are at much higher resolution, so coarser modalities may only have a single scalar value.

\subsubsection{Storage Format and Parallelism}
\label{ia:storage-parallelism}

\paragraph{Storage format.}
Per-modality tiles are written into per-modality subdirectories of the output root, with filenames keyed by product ID and window row/column. Each netCDF file stores per-band variables under the layer's named bands, the tile's CRS as a WKT string, and the per-axis pixel resolution as global attributes. Arrays are compressed into chunks using Bitshuffle+LZ4. Per-channel value-range clipping (e.g.\ albedo $\in [0,\,1]$, TiO\textsubscript{2} abundance $\in [0,\,100]\,\%$) is applied at write time using the limits declared in the per-modality value-range registry. 

\paragraph{Parallelism and resumability.}
Workers are dispatched through a process pool with a bounded in-flight queue, and Parquet flushes are performed in the main process to avoid many small writes. A bootstrap mode scans the existing output Parquet files for already-processed product IDs and skips them on restart. Pass 2 supports splitting work across array jobs and an optional local scratch staging directory with periodic synchronization to network storage.

\subsubsection{Records and Reproducibility}
\label{ia:records}

Together, the two tracks produce three classes of artifacts: the
per-modality netCDF tile files, the per-track tile catalogs
(\texttt{WAC\_LowRes.parquet} for the low-resolution track and
\texttt{NAC\_HighRes.parquet} for the high-resolution track),
and the split manifest. The
aggregate dataset sizes at the configured tile size are reported in
Table~\ref{tab:track-comparison}.

\subsubsection{Dataset Splits}
\label{ia:train-test-val-split}
This dataset is intended for large-scale model pre-training and evaluation on downstream applications. As such, it is split into three datasets
(\texttt{train}, \texttt{val}, and \texttt{test}) to prevent data leakage across tasks. Data that appears in the training and validation sets is used for pre-training,
whereas data in the test set is reserved for model evaluation and fine-tuning for downstream applications.
The data is randomly assigned to the training, validation, and test splits on a per-LTM basis. Each LTM zone appears in only one split.
Data in the LPS\_S and LPS\_N zones are spatially partitioned into the training, validation, and test sets, ensuring that overlapping tiles appear in only one split.
Zones and spatial partitions appear in the same split across the high-resolution and low-resolution tracks.
The data is split with a nominal 80/10/10 training/validation/test
target. Because entire zones are assigned wholesale, the achieved
proportions in the released artifact are 78.3\,/\,12.0\,/\,9.7\% for
the low-resolution track and 75.1\,/\,15.7\,/\,9.2\% for the
high-resolution track.

\begin{table}[!htbp]
\centering
\caption{Comparison of low-resolution and high-resolution track characteristics.}
\label{tab:track-comparison}
\small
\begin{tabular}{@{}lll@{}}
\toprule
\textbf{Component} & \textbf{Low-Resolution Track} & \textbf{High-Resolution Track} \\
\midrule
Optical Anchor & WAC VIS (100 m/px) & NAC (1 m/px) \\
Tile Resolution & 51.2 km $\times$ 51.2 km & 512 m $\times$ 512 m \\
Tile Size (pixels) & 512 $\times$ 512 & 512 $\times$ 512 \\
Rejection stride   & 120        & 120 \\
Pass 1 Extracts & VIS, UV, DTM (60m) & NAC only \\
Pass 2 Extracts & Static maps & NAC-derived DTMs + Static maps \\
Non-polar: Modalities requiring no NaN & VIS, UV, DTM (60m) & NAC \\
Polar: Modalities requiring any valid data & VIS only & NAC only \\
EDRs Processed & 54,080 WAC pairs & 1,107 NAC EDRs \\
Tiles Emitted & 963,609 & 1,000,113 \\
Total netCDF files (all modalities) & 21,473,477 & 21,762,559 \\
\bottomrule
\end{tabular}
\end{table}

\begin{table*}[!htbp]
\centering
\caption{Per-modality coverage for the low-resolution and high-resolution tracks.
Counts indicate the number of emitted netCDF files for each modality, and Avg Null Pct
is the average percent of null pixels per tile. Dataset key indicates the parquet column associated with this modality. Modality resolution for both the high-resolution and low-resolution datasets after resampling procedure also included.}
\label{tab:modality-coverage-and-resolution}
\small
\resizebox{\textwidth}{!}{%
\begin{tabular}{@{}L{3.0cm}L{3.2cm}ccccccc@{}}
\toprule
\textbf{Modality} &
\textbf{Dataset key} &
\textbf{\makecell{Native Resolution\\(m)}} &
\textbf{\makecell{Low-resolution\\count}} &
\textbf{\makecell{Low-resolution\\avg. null (\%)}} &
\textbf{\makecell{Low-resolution\\ snapped resolution (m)}} &
\textbf{\makecell{High-resolution\\count}} &
\textbf{\makecell{High-resolution\\avg. null (\%)}} &
\textbf{\makecell{High-resolution\\ snapped resolution (m)}}\\
\midrule
NAC Panchromatic   & 
    NAC           & 1.0      & --         & --      & --         & 1,000,113 & 0.14\%  & 1.0    \\
\makecell[l]{WAC VIS\\(\SIrange{415}{689}{\nano\meter})} &
    WAC\_VIS      & 100.0    & 963,609    & 0.24\%  & 100.0      & --        & --      & --     \\
\makecell[l]{WAC UV\\(\SIrange{321}{360}{\nano\meter})} & 
    WAC\_UV       & 500.0    & 963,609    & 0.26\%  & 501.96     &--         & --      & --     \\
Topography (SLDEM2015) &
    \makecell[l]{DTM\_60M (high-res)\\DTM (low-res)}      & 60.0     & 963,609    & 0.00\%  & 60.02      & 1,000,113 & 0.00\%  & 56.89  \\
NAC-stereo DTM & 
    DTM\_3M       & 3.0      & --         & --      & --         & 999,824   & 0.12\%  & 2.99   \\
\makecell[l]{Topography (SLDEM2015)\\Slope}  &
    \makecell[l]{SLOPE\_60M (high-res)\\SLOPE (low-res)}      & 60.0     & 963,609    & 0.00\%  & 60.02      & 1,000,113        & 0.00\%      & 56.89  \\
NAC-stereo DTM (Slope) &
    SLOPE\_3M   & 3.0      & --         & --      & --         & 999808 & 0.00\%  & 2.99   \\
\makecell[l]{Topography (SLDEM2015)\\Azimuth} &
    \makecell[l]{ASPECT\_60M (high-res)\\ASPECT (low-res)}      & 60.0     & 963,609    & 0.00\%  & 60.02      & 1,000,113 & 0.00\%  & 56.89  \\
NAC-stereo DTM (Aspect) &
    ASPECT\_3M    & 3.0      & --         & --      & --         & 999,808   & 0.12\%  & 2.99   \\
Unified Geologic Map &
    GEOMAP       & 60.0     & 963,609    & 0.16\%  & 60.02      & 1,000,113 & 0.00\%  & 56.89  \\
Diviner $T_{\rm reg}$ Anomaly &
    TREG         & 240.0    & 925,387    & 5.11\%  & 240.38     & 991,337   & 0.89\%  & 256.0  \\
LOLA Roughness &
    ROUGHNESS    & 1,000.0  & 962,886    & 15.06\% & 1,0003.92  & 867,020   & 13.31\% & 512.0  \\
Mini-RF Radar circular polarization ratio &
    MINIRF\_CPR  & 90.0     & 940,383    & 41.39\% & 89.98      & 644,591   & 37.58\% & 85.33  \\
Mini-RF Radar reflectivity &
    MINIRF\_S1   & 90.0     & 940,723    & 40.47\% & 89.98      & 649,624   & 36.87\% & 85.33  \\
WAC TiO\textsubscript{2} &
    TIO2         & 400.0    & 925,622    & 4.42\%  & 400.0      & 992,113   & 0.80\%  & 512.0  \\
WAC Morphologic Mosaic &
    WAC\_MOSAIC  & 100.0    & 963,609    & 0.00\%  & 100.0      & 1,000,113 & 0.00\%  & 102.4  \\
Kaguya MI Mineralogy &
    MI\_MINERALOGY & 60.0     & 732,004    & 25.99\% & 60.02      & 947,764   & 5.47\%  & 56.89  \\
Kaguya MI Norm.\ Reflectance  &
    MI\_NORM\_REF & 60.0     & 892,684    & 9.16\%  & 60.02      & 977,324   & 2.51\%  & 56.89  \\
Kaguya MI Space Weathering &
    SW\_FE      & 1,000.0  & 843,199    & 13.99\% & 1,0003.92  & 971,424   & 2.87\%  & 512.0  \\
WAC Norm.\ Reflectance &
    WAC\_NORM\_REF & 500.0    & 843,262    & 13.22\% & 501.96     & 985,729   & 1.44\%  & 512.0  \\
WAC \SI{643}{\nano\meter} HR &
    WAC\_NR\_643\_HR  & 100.0    & 963,609    & 0.18\%  & 100.0      & 1,000,099 & 0.02\%  & 102.4  \\
Kaguya SP Polar Mineralogy &
    SP\_MINERALOGY & 1,000.0  & 6,871      & 99.33\% & 10,0003.92  & 6,055     & 64.48\% & 512.0  \\
LOLA Permantly shadowed regions &
    PSR  & 20.0     & 6,871      & 99.29\% & 20.0       & 6,072     & 64.48\% & 19.69  \\
LOLA Avg.\ Sun Illumination &
    AVG\_ILLUM    & 120.0    & 6,871      & 99.29\% & 119.91     & 6,072     & 64.48\% & 128.0  \\
LOLA \SI{1064}{\nano\meter} Albedo &
    ALBEDO   & 1,000.0  & 6,871      & 99.29\% & 1,0003.92  & 6,072     & 64.48\% & 512.0  \\
Diviner Ice Stability Depth &
    DICE     & 240.0    & 6,871      & 99.46\% & 240.38     & 4,782     & 64.63\% & 256.0  \\
Diviner $T_{\rm bol}$ polar (24-phase, summer/winter) &
    TBOL\_POLES         & 240.0    & 6,871      & 99.51\% & 240.38     & 6,072     & 99.62\% & 256.0  \\
Diviner $T_{\rm bol}$ (24-phase) &
    TBOL    & 15,000.0 & 963,609    & 0.00\%  & 17,066.67  & 1,000,113 & 0.00\%  & 512.0  \\
GRAIL Free-air Gravity &
    GRAVITY & 20,000.0 & 963,609 & 0.00\%  & 17,066.67  & 1,000,113 & 0.00\%  & 512.0  \\
LP Hydrogen Abundance &
    HYDROGEN  & 15,000.0 & 6,871      & 99.29\% & 17,066.67  & 6,072     & 64.48\% & 512.0  \\
Diviner H-parameter &
    HPAR   & 240.0    & 930,535    & 4.60\%  & 240.38     & 344,747   & 1.15\%  & 256.0  \\
Diviner Rock Abundance & 
    ROCK\_ABUNDANCE  & 240.0    & 925,387    & 5.11\%  & 240.38     & 349,133   & 0.39\%  & 256.0  \\
Associated metadata (illumination, location, etc.) &
    METADATA   & --       & 963,609    & 0.00\%  & --         & 1,000,113 & 0.00\%  & --     \\
Robbins crater labels \cite{robbins2019new} &
    CRATERS   & --       & 963,609    & 0.00\%  & 512.0      & --        & --      & --     \\
\bottomrule
\end{tabular}%
}
\end{table*}

\subsection{Application Benchmark Datasets}
\label{sec:benchmarks}
To provide a reusable testing environment and to connect benchmark design to community priorities, we define initial \dataset tasks and associated datasets aligned with three broad lunar science themes: impact processes, volcanic history, and polar volatiles. These themes motivate benchmark tasks that span detection/localization, segmentation, and classification/regression, and are designed to exercise both single-modality and multimodal inputs under the illumination and resolution heterogeneity intrinsic to lunar observations. These tasks are by no means exhaustive, but rather are meant to be illustrative of common ML applications in lunar science and serve as a starting point for more advanced applications currently under development. For each task, we release a dataset, including inputs, labels, splits, and evaluation metrics, that can be used to enable reproducible comparison of methods. The benchmarks are built on the same processed inputs as the pre-training corpus. Where benchmark tiles are drawn from the corpus itself, they are taken from its held-out test split (Section~\ref{ia:train-test-val-split}) so that benchmark evaluation data are never seen during pre-training.

\subsubsection{Impact Processes}    

Impact craters are the most ubiquitous landforms on the Moon and provide a natural benchmark for ML because they span orders of magnitude in scale, occur on nearly all terrain types, and support standardized evaluation for detection, segmentation, and size–frequency analyses. Crater catalogs also underpin a wide range of downstream lunar investigations, from relative-age dating to geologic mapping and hazard-context assessments, making them a high-value target for reproducible benchmarking.

\paragraph{Robbins Crater Catalog} 

Our primary crater dataset is one of the most widely-used lunar crater catalogs \cite{robbins2019new}, which we refer to as the "Robbins catalog" (or dataset/database). It is a manually-compiled global database of lunar impact craters intended to be a near-complete census of diameters $\geq1 -2$ km \cite{robbins2019new}. The catalog contains 2,033,574 mapped craters in total, including 1,296,879 craters $\geq 1$~km, $\sim83,000 \geq 5$~km, and 6,972 $\geq $~20 km \cite{robbins2019new}. The database was produced through repeated full-Moon manual searches conducted primarily using LROC WAC global mosaics optimized for morphological interpretation and individual WAC images, with crater rims digitized in GIS and fit to standardized crater centers and diameters. 
Independent topographic products were used as a cross-check where helpful, particularly under challenging illumination/high-latitude conditions \cite{robbins2019new}. To construct an object detection benchmark from this catalog, we selected 1,000 WAC visible tiles from the test split of the \dataset low-resolution track (Section~\ref{ia:train-test-val-split}), so they were not seen by the model during pre-training, restricting the selection to images with incidence angles between 60\degree{} and 80\degree{} to obtain more favorable illumination conditions of the craters. For each selected tile, the Robbins catalog craters falling within the tile's spatial bounds were converted to bounding boxes in the format of \textit{x\_min}, \textit{y\_min}, \textit{width}, \textit{height} and packaged, together with the corresponding DTM tiles, as COCO annotations. Summary statistics for these images in Table\ref{tab:wac_robbins_statitstics} show that there are on average \textasciitilde{}114 craters per image.

\begin{table}[H]
\centering
\small
\caption{WAC Robbins COCO dataset splits with per-split image and annotation counts, mean craters per image, median crater size ($\sqrt{\text{area}}$ in px), and mean illumination geometry.}
\label{tab:wac_robbins_statitstics}
\begin{tabular}{lrrrrrr}
\toprule
\textbf{Split} & \textbf{Images} & \textbf{Annotations} & \makecell{\textbf{Craters}\\\textbf{/ img}} & \makecell{\textbf{Median}\\$\sqrt{\text{area}}$} & \makecell{\textbf{Incidence}\\(\degree)} & \makecell{\textbf{Emission}\\(\degree)} \\
\midrule
Train & 800 & 91,516 & 114.4 & 16.5 & 72.2 & 1.81 \\
Val   & 100 & 11,429 & 114.3 & 16.5 & 71.1 & 1.55 \\
Test  & 100 & 11,384 & 113.8 & 17.0 & 72.7 & 1.39 \\
\bottomrule
\end{tabular}
\end{table}

\paragraph{Additional Hand-Labeled Images} 

\begin{figure}
    \centering
    \includegraphics[width=\linewidth]{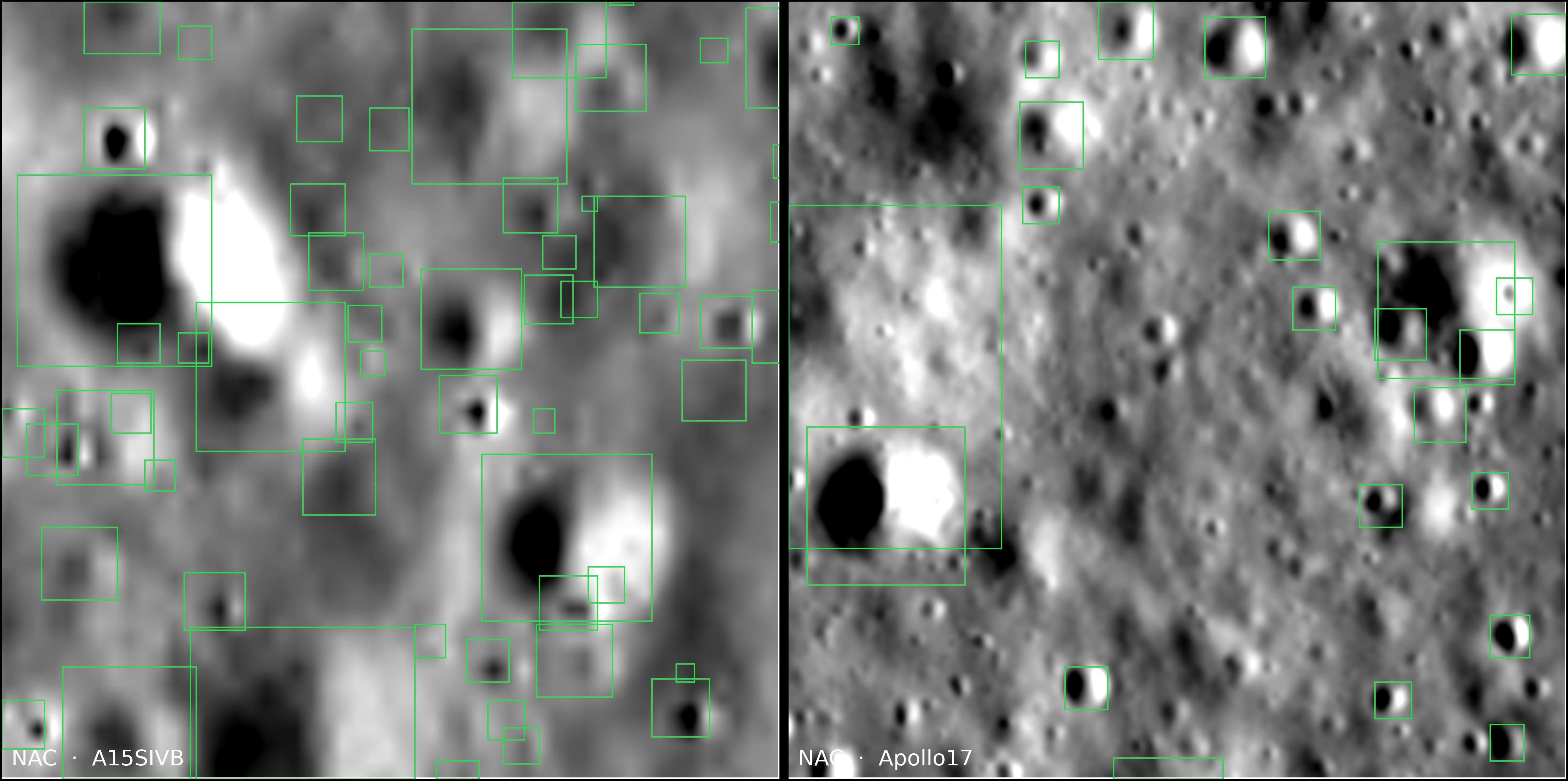}
    \caption{Hand-labeled crater annotations on two NAC study areas, Apollo 15 S-IVB (left) and Apollo 17 (right). Green boxes denote annotated crater instances. Apollo 15 S-IVB was labeled at 5 m/pixel resolution, causing it to appear blurrier than the Apollo 17 image which was labeled at a higher resolution.}
    \label{fig:nac_handlabeled}
\end{figure}

In addition to the Robbins’ catalog, we constructed a higher resolution, manually annotated crater dataset focused on regions surrounding the NAC\_PHO sites. We selected LROC NAC orthophotos at 0.8 to 5 m/pixel acquired under favorable relief-enhancing illumination (incidence angle \textasciitilde{}50–80°) for six NAC\_PHO sites (Apollo 15 S-IVB, Apollo 17, Highlands Photometric, King Ejecta, March 17 Impact Crater, and Reiner Gamma). Imagery was overlaid with the corresponding NAC DTMs and inspected in QGIS. For each site, we defined 5–9 study boxes of 1,024 × 1,024 m (35 in total) chosen to sample varied terrain and a broad range of crater diameters. Craters were manually digitized as circles using OpenCraterTool \citep{Heyer2023}, with two annotation passes per box merged and de-duplicated (IoU > 0.5). Each released tile carries a \texttt{label\_set} flag: \textit{standard} tiles are exhaustively labeled at all crater diameters, while \textit{large} tiles are labeled only for craters at least 10 m in diameter, which should be accounted for when evaluating detections on those tiles. After labeling, each box was tiled into 256 x 256 pixel patches at the native image resolution (0.8–5 m/pixel) using an even-overlap grid and packaged into COCO format. In total, there are 408 image tiles with 57,611 crater annotations covering 49,106 unique craters. A crater may appear in more than one tile because adjacent tiles within a box overlap. The train/val/test splits (289/56/63 tiles) are assigned at the box level, so overlapping tiles never cross splits and there is no data leakage. Boxes labeled at 5 m/pixel resolution, such as Apollo 15 S-IVB, appear blurrier than those labeled at a higher resolution (Figure~\ref{fig:nac_handlabeled}).

\subsubsection{Volcanic History} 

Irregular mare patches (IMPs) provide a surface feature benchmark that is complementary to impact features because they are rare, morphologically diverse, and often defined by subtle boundary expression rather than strong, ubiquitous signatures. IMPs span a broad range of sizes, from tens of meters to \textasciitilde{}5 km \cite{hargitai2025clusters}, and their diffuse margins make dense, pixel-level supervision necessary for reliable delineation. \dataset therefore frames IMP mapping as a binary segmentation task. The benchmark comprises 130 image tiles manually labeled with pixel-level IMP segmentation masks, split into 100 training, 20 validation, and 10 test tiles, to support standardized evaluation of segmentation performance across illumination conditions and terrain settings. 

\subsubsection{Polar Volatiles}

To support polar-volatiles benchmarking, \dataset incorporates a polar map stack derived from the workflow in \cite{coyan2025prospectivity}, who developed a lunar ice prospectivity model using a high-resolution DTM and a numerical thermal model to generate a set of geospatial inputs that are plausibly predictive of near-surface water ice \cite{coyan2025prospectivity}. In \dataset, we package nine static layers within $10\degree$ latitude of each pole at 240 m/pixel sampling.
The benchmark is framed as regression: the eight evidential layers serve as inputs and the continuous (0--1) ice-prospectivity layer is the target, with 162 patches spanning both poles split into 108 training, 24 validation, and 25 test patches. The same layers are also designed to function as standardized inputs for related tasks, such as polar terrain classification and ice-prospectivity prediction under extreme illumination regimes.

The nine polar layers are: (1) ice stability depth, (2) maximum surface temperature, (3) PSRs, (4) slope, (5) curvature, (6) aspect (provided as a sine--cosine encoding, with an alternative cosine-only variant), (7) distance to PSRs, (8) PSR density, and (9) polar ice prospectivity \cite{coyan2025prospectivity}. These layers intentionally combine (i) thermophysical constraints (ice stability depth and maximum temperature), (ii) illumination state (PSRs), and (iii) terrain controls that modulate both illumination and regolith transport (slope, curvature, aspect), with additional context features that capture spatial relationships to cold-trap environments (distance to PSRs and PSR density) \cite{coyan2025prospectivity}. The final prospectivity map represents a continuous estimate of where water ice is most likely within the upper \textasciitilde{}1 m of regolith, given assumed relationships between ice concentration and the evidential layers \cite{coyan2025prospectivity}. Critically, it is designed as a tunable product whose weights can evolve as new in situ constraints become available.

Topographic derivatives (slope, aspect, curvature) were computed from the DTM, and thermal layers (maximum temperature and ice stability depth) derived from a numerical thermal model calibrated to Diviner observations \cite{coyan2025prospectivity}. PSRs are treated as an explicit evidence layer because permanent shadow can be independently relevant beyond its correlation with modeled temperatures. PSR determination is performed via horizon/visibility modeling over long time windows (i.e., evaluating whether any portion of the solar disk becomes visible above the local horizon over repeated cycles) \cite{coyan2025prospectivity}. Finally, two proximity/context layers are constructed to capture plausible transport/mixing or “halo” effects around PSRs: distance to PSRs (nearest-neighbor distance) and PSR density (local areal density of PSRs within a fixed-radius neighborhood), which provide interpretable features for ML models attempting to generalize ice-related patterns beyond PSR interiors \cite{coyan2025prospectivity}.

\rowcolors{2}{}{}
\begin{table*}[!htbp]
\centering
\small
\caption{Overview of \dataset benchmark datasets separated by task type.}
\label{tab:downstream-bench-overview}
\setlength{\tabcolsep}{4pt}
\renewcommand{\arraystretch}{1.05}
\resizebox{\textwidth}{!}{%
\begin{tabular}{@{}l l c c c r r r@{}}
\toprule
\textbf{Name} &
\textbf{Science Theme} &
\textbf{Image Size (px)} &
\textbf{Resolution (m/px)} &
\textbf{Target} &
\textbf{Train} &
\textbf{Val} &
\textbf{Test} \\
\midrule
\multicolumn{8}{@{}l}{\textbf{Object Detection}} \\
\midrule
Robbins craters & Impact processes & 512 $\times$ 512 & 100 & 1 class & 800 & 100 & 100 \\
NAC hand labeled & Impact processes & 256 $\times$ 256 &  0.8--5 & 1 class & 289 & 56 & 63\\
\midrule
\multicolumn{8}{@{}l}{\textbf{Segmentation}} \\
\midrule
Irregular Mare Patches & Volcanic History & 256 $\times$ 256 & 1 & 1 class & 100 & 20 & 10 \\
\midrule
\multicolumn{8}{@{}l}{\textbf{Regression}} \\
\midrule
Lunar ice prospectivity & Polar Volatiles & 256 $\times$ 256 & 240 & 0--1 & 108 & 24 & 25 \\
\bottomrule
\end{tabular}
}
\end{table*}

Together, this multi-layer polar stack provides a compact, physically motivated input suite for benchmarking models that aim to predict and explain the likelihood of the presence of ice throughout the lunar polar environment. In \dataset, these layers enable standardized evaluation of approaches ranging from simple baselines to multimodal models under consistent spatial extent, resolution, and polar projection conventions.

These benchmark datasets are packaged using a common construction pipeline that standardizes projections, tiling, metadata, and train/validation/test splits across modalities and regions. Table \ref{tab:downstream-bench-overview} shows an overview of the benchmark datasets available in \dataset. The next section describes how the released pre-training and benchmark artifacts are hosted and organized on disk.

\section{Data Records}
\label{sec:data-records}

This section describes how the released artifacts are hosted, organized
on disk, and consumed. It complements the construction descriptions in
Sections~\ref{sec:construction} and \ref{sec:benchmarks}, which explained how each artifact was produced. This section describes the
form in which a downstream user encounters it.

\subsection{Repository and access}

The downstream benchmark datasets are hosted on Hugging Face in the
\href{https://huggingface.co/collections/nasa-ibm-ai4science/sombench}{\texttt{nasa-ibm-ai4science/sombench}}
collection. 


\subsection{On-disk organization}

\dataset is distributed as per-modality netCDF tile directories
indexed by Parquet catalogs, written to a common root directory.

\paragraph{Artifact layout.}
The pipeline
(Section~\ref{sec:image_anchored_pipeline}) emits per-modality
sub-directories of the output root. Each sub-directory holds netCDF
tiles named after the source EDR product id and the sliding-window
row/column (\texttt{\{pid\}\_r\{row\}\_c\{col\}.nc}). Two Parquet
catalogs record tile-level metadata:
\texttt{WAC\_LowRes.parquet} for the low-resolution track (anchored by WAC)
 and \texttt{NAC\_HighRes.parquet} for the high-resolution
 track (anchored by NAC). 

\paragraph{Aggregate scale.} 
At the released configuration (Table~\ref{tab:track-comparison}), the image-anchored artifact contains 963{,}609 multimodal
tiles from 54{,}080 WAC EDR pairs in the
low-resolution track with 29 modalities and 1{,}000{,}113 NAC tiles from
1{,}107 NAC EDRs in the high-resolution track with
31 modalities attached, with 42.48\,TB total size.

\subsection{Tile file format}

Every tile is a self-describing netCDF4 file written through
\texttt{h5netcdf} with Bitshuffle + LZ4 chunk compression
(\texttt{hdf5plugin.Bitshuffle(cname="lz4")}; one chunk per variable,
chunk shape equal to the full array shape). Each tile carries:
\begin{itemize}[nosep,leftmargin=1.2em]
  \item \textbf{Per-band variables} with explicit names when there are multiple channels, otherwise label as \texttt{band\_data}. Topographic layers use physical
        names (e.g., \texttt{Aspect\_sin}, \texttt{Aspect\_cos}). Spectral layers
        use band wavelength in nanometers (\texttt{415},
        \texttt{566}, \texttt{604}, \texttt{643}, \texttt{689} for
        WAC VIS, and \texttt{321}, \texttt{360} for WAC UV).
        Compositional layers use mineral names
        (\texttt{olivine}, \texttt{plagioclase}, \texttt{hcp},
        \texttt{lcp}, \texttt{feo}, \texttt{omat}). The full list of
        per-modality band names is given in
        Table~\ref{tab:modality-coverage-and-resolution}.
  \item \textbf{Coordinate vectors.} Per-axis pixel-center $(x, y)$
        coordinates in the tile's CRS, stored as 1D \texttt{x} and
        \texttt{y} variables.
  \item \textbf{Global attributes.} The full CRS as a WKT string
        (\texttt{crs}), per-axis pixel resolution
        (\texttt{pix\_res\_x}, \texttt{pix\_res\_y}), and a
        comma-separated list of band names (\texttt{band\_names}).
\end{itemize}
Per-channel value-range clipping is applied at write time using
limits declared in the per-modality value-range registry (e.g.\
slope $\in [0\degree, 90\degree]$, normalized reflectance
$\in [0,\,1]$, MI mineralogy $\in [0,\,100]\,\%$, elevation
$\in [-9500,\,10800]\,$m), while NaN encodes a genuine data gap and is
preserved through clipping. Categorical layers (the unified geologic
map) are exempt from clipping so class indices remain unchanged.

\subsection{Parquet catalog schemas}

Each row in \texttt{WAC\_LowRes.parquet} or
\texttt{NAC\_HighRes.parquet} corresponds to a single
sliding-window tile and carries: \texttt{PRODUCT\_ID} (the source
EDR id), \texttt{ROW} and \texttt{COL} (window pixel offset in the
EDR canvas), \texttt{LTM\_CODE} (LTM zone or LPS cap codes), per-modality
\texttt{\{MODALITY\}\_TILE} relative paths and
\texttt{\{MODALITY\}\_FRACTION\_NULL} NaN fractions, the four corner
lon/lat coordinates and projected bounds, and EDR-derived
viewing-geometry columns (incidence, emission, and phase angles and
sub-solar/sub-spacecraft geometry) extracted from the LROC EDR
labels. The parquet catalogs have a \texttt{DATASET} column that indicates to which split (\texttt{train/val/test}) the tile was assigned based on the logic outlined in \ref{ia:train-test-val-split}.  The \texttt{ALLOW\_NANS\_OPTICAL\_DTM\_SLOPE\_ASPECT} column flags if there are NaNs in the optical (WAC vis/uv/NAC) or the terrain modalities (DTM/slope/aspect).

\subsection{Benchmark dataset records}

Each application benchmark (Section~\ref{sec:benchmarks}) is released as a self-contained artifact alongside the pre-training corpus, with its train/validation/test split materialized on disk:
\begin{itemize}[nosep,leftmargin=1.2em]
  \item \textbf{Robbins WAC craters:} 1,000 WAC visible tiles (512 $\times$ 512 pixels at 100 m/pixel) with the corresponding DTM tiles and per-split COCO-format annotation files derived from the Robbins catalog \citep{robbins2019new}.
  \item \textbf{NAC hand-labeled craters:} 408 NAC orthophoto tiles (256 $\times$ 256 pixels at 0.8--4 m/pixel) with per-split COCO-format annotation files and the per-tile \texttt{label\_set} flag (\textit{standard} vs.\ \textit{large}) described in Section~\ref{sec:benchmarks}.
  \item \textbf{Irregular Mare Patches:} 130 image tiles (256 $\times$ 256 pixels) paired with pixel-level binary segmentation masks.
  \item \textbf{Ice prospectivity:} 162 polar patches (256 $\times$ 256 pixels at 240 m/pixel), each stacking the eight evidential input layers and the target prospectivity layer.
\end{itemize}

\section{Technical Validation}
\label{sec:tech-validation}
\subsection{Baseline Models}

To verify that each \dataset benchmark task is learnable from the released inputs and that the benchmark datasets are compatible with published, off-the-shelf ML models, we train two representative supervised baselines on all benchmark datasets: ResNet-50 \cite{he2016deep}, initialized from ImageNet-pretrained weights, as a standard convolutional reference, and SwinV2-Base \cite{liu2022swin}, an ImageNet-pretrained hierarchical vision transformer. Together they cover the two dominant architecture families in modern computer vision. The intent is to demonstrate compatibility and learnability, not to rank models. Each backbone is paired with a task-appropriate framework for each benchmark, described below. All baselines are implemented with TerraTorch \cite{gomes2025terratorch} and trained with the AdamW optimizer under fixed seeds and deterministic augmentation. For each run, the checkpoint that performs best on the validation split under the task metric described below is retained for test-set evaluation.

\subsubsection{Crater Detection}
\begin{figure}[!htbp]
    \centering
    \includegraphics[width=\linewidth]{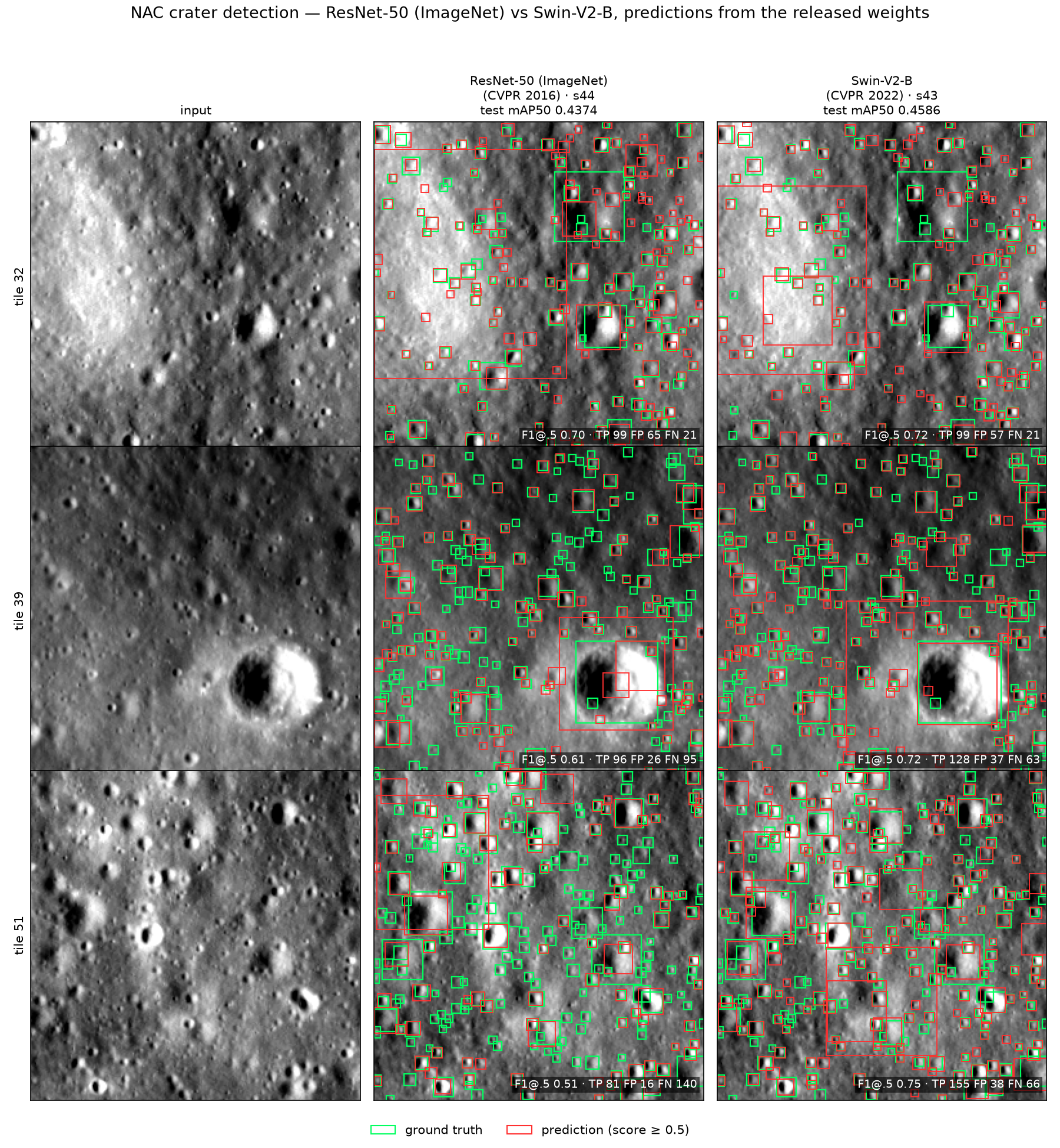}
    \caption{Qualitative crater detection results on three test tiles of the NAC hand-labeled benchmark. Columns show the input tile and predictions from SwinV2-B, and ResNet-50 (ImageNet), using the released weights. Green boxes denote ground-truth annotations and red boxes model predictions with confidence $\geq$ 0.5. Per-panel insets report F1@0.5 and true-positive, false-positive, and false-negative counts.}
    \label{fig:nac-craters-qual}
\end{figure}

For the WAC Robbins catalog and NAC hand-labeled crater benchmarks, each backbone is embedded in a Faster R-CNN detection framework \cite{ren2015faster}. Models are trained for 100 epochs and selected on validation mean Average Precision (mAP). To probe sample efficiency on the larger WAC benchmark, both models are additionally trained in a half-training-data configuration. Table~\ref{tab:downstream-crater-nac} reports test-set bounding-box mAP on the NAC hand-labeled benchmark, averaged over five random seeds. Absolute scores are low for both models, particularly at the stricter IoU thresholds, reflecting the difficulty of localizing dense fields of small craters near the resolution limit of the imagery. Figure~\ref{fig:nac-craters-qual} shows qualitative predictions on three NAC test tiles.

\begin{table}[!htbp]
\centering
\caption{Crater detection on the NAC hand-labeled benchmark (meter-scale LROC NAC imagery). Test-set bounding-box mean average precision: mAP (IoU 0.50:0.95), AP@50, and AP@75. Values are mean $\pm$ standard deviation over five random seeds.}
\label{tab:downstream-crater-nac}
\begin{tabular}{lccc}
\toprule
Model & mAP & AP@50 & AP@75 \\
\midrule
ResNet-50 (ImageNet)   & 0.1411 $\pm$ 0.0036 & 0.4374 $\pm$ 0.0196 & 0.0895 $\pm$ 0.0055 \\
SwinV2-B (ImageNet)    & 0.1552 $\pm$ 0.0086 & 0.4586 $\pm$ 0.0308 & 0.1090 $\pm$ 0.0146 \\
\bottomrule
\end{tabular}
\end{table}

\begin{figure}[!htbp]
    \centering
    \includegraphics[width=\linewidth]{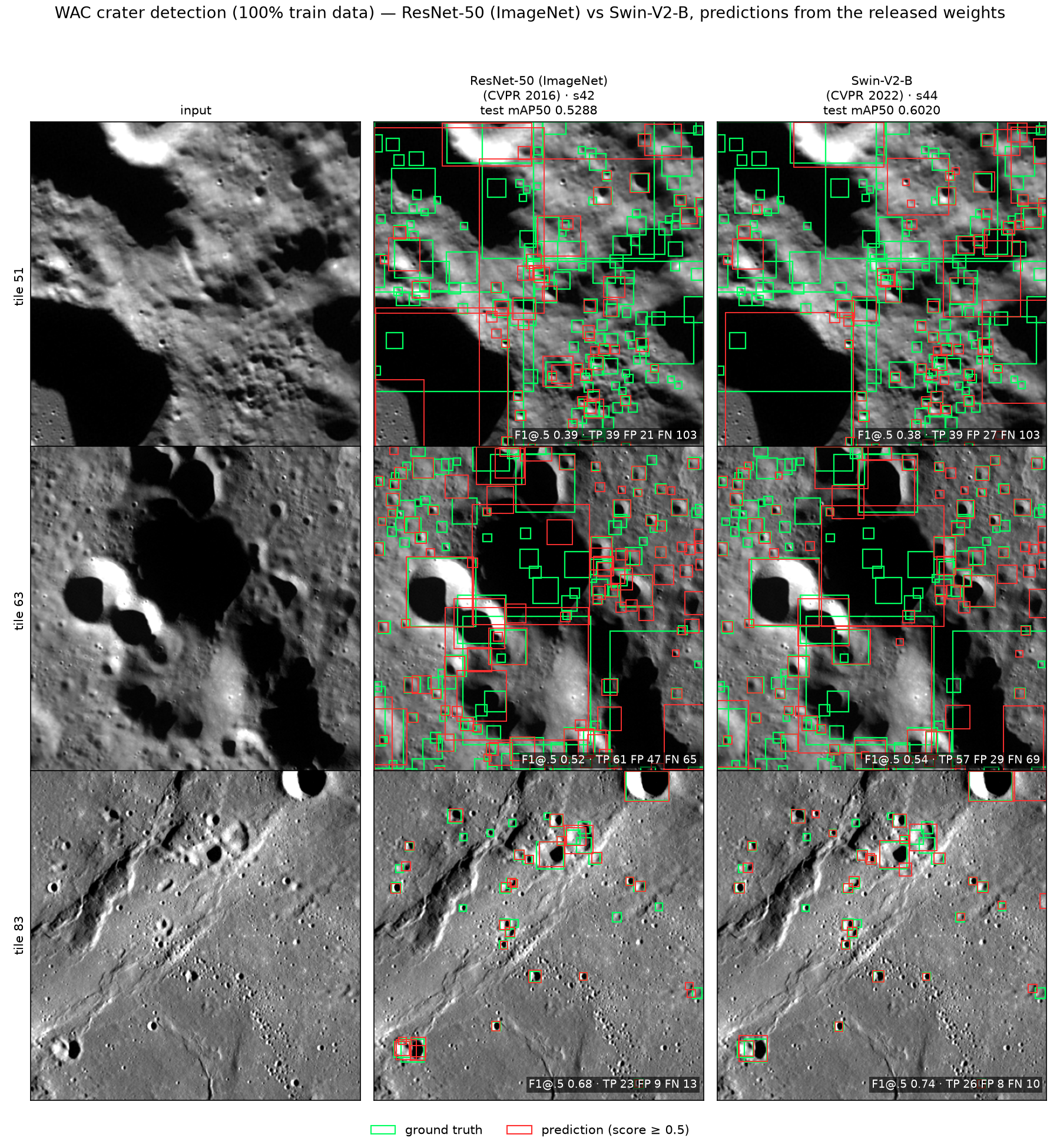}
    \caption{Qualitative crater detection results on three test tiles of the WAC Robbins catalog benchmark at 100\% training data. Columns show the input tile and predictions from ResNet-50 (ImageNet) and SwinV2-B, using the released weights. Green boxes denote ground-truth annotations and red boxes model predictions with confidence $\geq$ 0.5. Per-panel insets report F1@0.5 and true-positive, false-positive, and false-negative counts.}
    \label{fig:wac-craters-full-qual}
\end{figure}

Table~\ref{tab:downstream-crater-wac} reports the WAC Robbins catalog benchmark at 100\% and 50\% training data. Both models transfer to the coarser, context-scale WAC imagery, and halving the training data lowers scores only modestly (e.g., SwinV2-Base mAP drops from 0.2420 to 0.2313). Figure~\ref{fig:wac-craters-full-qual} shows qualitative predictions on three WAC test tiles.

\begin{table}[!htbp]
\centering
\caption{Crater detection on the WAC Robbins catalog benchmark (context-scale LROC WAC imagery) as a function of training-data fraction (left: 50\% training data, right: 100\%). Test-set bounding-box mean average precision: mAP (IoU 0.50:0.95), AP@50, and AP@75. Values are mean $\pm$ standard deviation over five random seeds.}
\label{tab:downstream-crater-wac}
\setlength{\tabcolsep}{4pt}
\resizebox{\textwidth}{!}{%
\begin{tabular}{lcccccc}
\toprule
& \multicolumn{3}{c}{\textbf{50\% training data}} & \multicolumn{3}{c}{\textbf{100\% training data}} \\
\cmidrule(lr){2-4} \cmidrule(lr){5-7}
Model & mAP & AP@50 & AP@75 & mAP & AP@50 & AP@75 \\
\midrule
ResNet-50 (ImageNet)   & 0.1993 $\pm$ 0.0007 & 0.4960 $\pm$ 0.0036 & 0.1572 $\pm$ 0.0038 & 0.2148 $\pm$ 0.0018 & 0.5288 $\pm$ 0.0055 & 0.1762 $\pm$ 0.0038 \\
SwinV2-B (ImageNet)    & 0.2313 $\pm$ 0.0027 & 0.5849 $\pm$ 0.0072 & 0.1862 $\pm$ 0.0041 & 0.2420 $\pm$ 0.0047 & 0.6020 $\pm$ 0.0061 & 0.2028 $\pm$ 0.0086 \\
\bottomrule
\end{tabular}%
}
\end{table}

\subsubsection{IMP Segmentation}
\begin{figure}[!htbp]
    \centering
    \includegraphics[width=\linewidth]{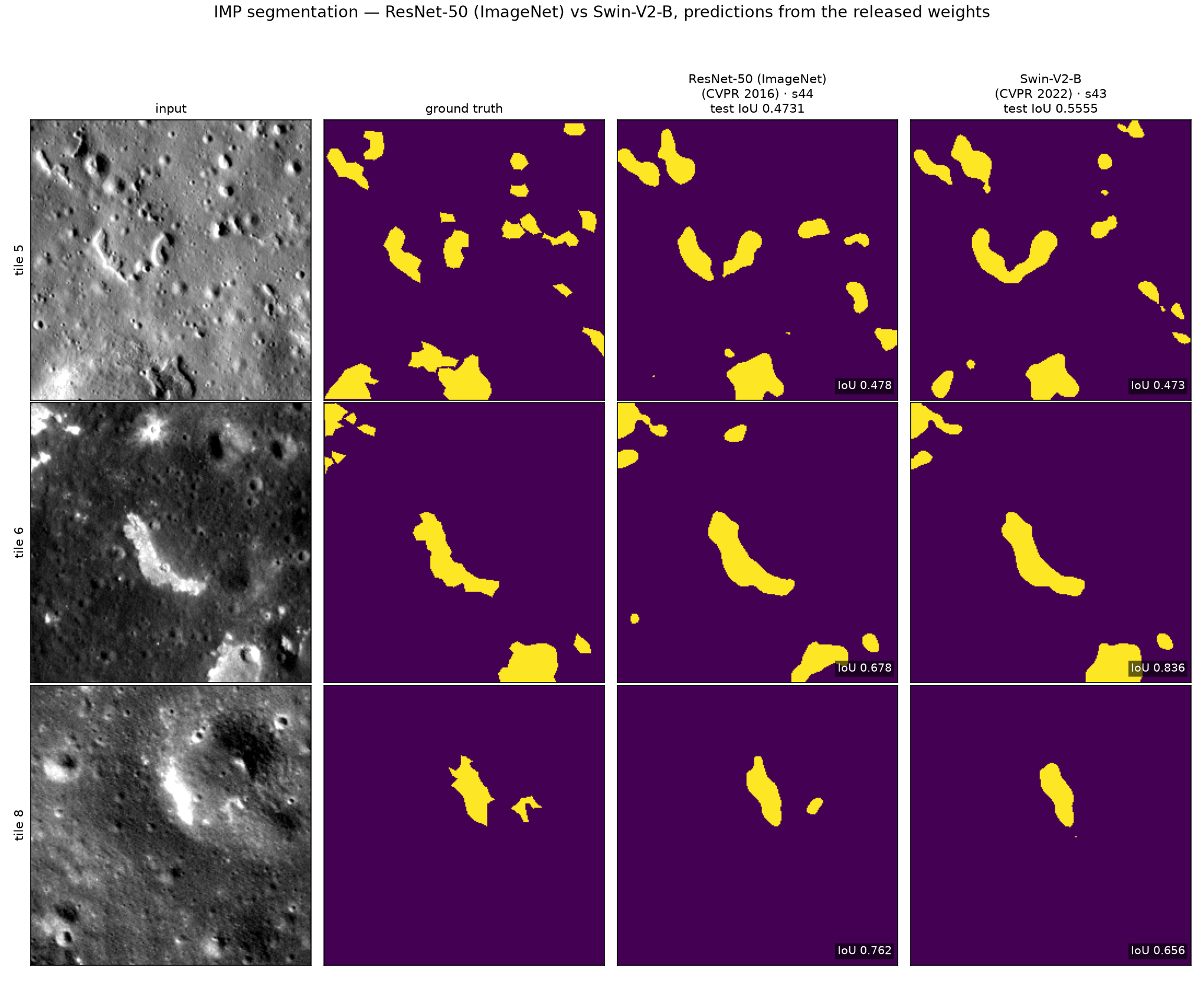}
    \caption{Qualitative IMP segmentation results on three test tiles. Columns show the input tile, the ground-truth mask, and predictions from ResNet-50 (ImageNet) and SwinV2-B, using the released weights. Yellow denotes the positive (IMP) class, and per-panel insets report the per-tile IoU.}
    \label{fig:imp-qual}
\end{figure}

The IMP benchmark is evaluated as binary semantic segmentation, pairing each backbone with a U-Net decoder \cite{ronneberger2015u}. Models are trained for 600 epochs with a combined Dice and cross-entropy loss and selected based on validation F1 score. Table~\ref{tab:downstream-imp} reports test-set intersection-over-union and F1 on the positive (IMP) class, averaged over five random seeds. Both models learn to delineate IMPs from the 100 training tiles. Figure~\ref{fig:imp-qual} shows qualitative predictions on three test tiles.

\begin{table}[!htbp]
\centering
\caption{Irregular Mare Patch (IMP) segmentation. Since IMP delineation is a binary dense-prediction task, we report test-set metrics on the positive (IMP) class only: intersection-over-union (IoU$_1$) and F1 score (F1$_1$). Values are mean $\pm$ standard deviation over five random seeds.}
\label{tab:downstream-imp}
\setlength{\tabcolsep}{6pt}
\begin{tabular}{l cc}
\toprule
Model & IoU$_1$ & F1$_1$ \\
\midrule
ResNet-50 (ImageNet)   & 0.4731 $\pm$ 0.0114 & 0.6423 $\pm$ 0.0105 \\
SwinV2-B (ImageNet)    & 0.5555 $\pm$ 0.0159 & 0.7141 $\pm$ 0.0132 \\
\bottomrule
\end{tabular}
\end{table}

\subsubsection{Ice Prospectivity Regression}
\begin{figure}[!htbp]
    \centering
    \includegraphics[width=\linewidth]{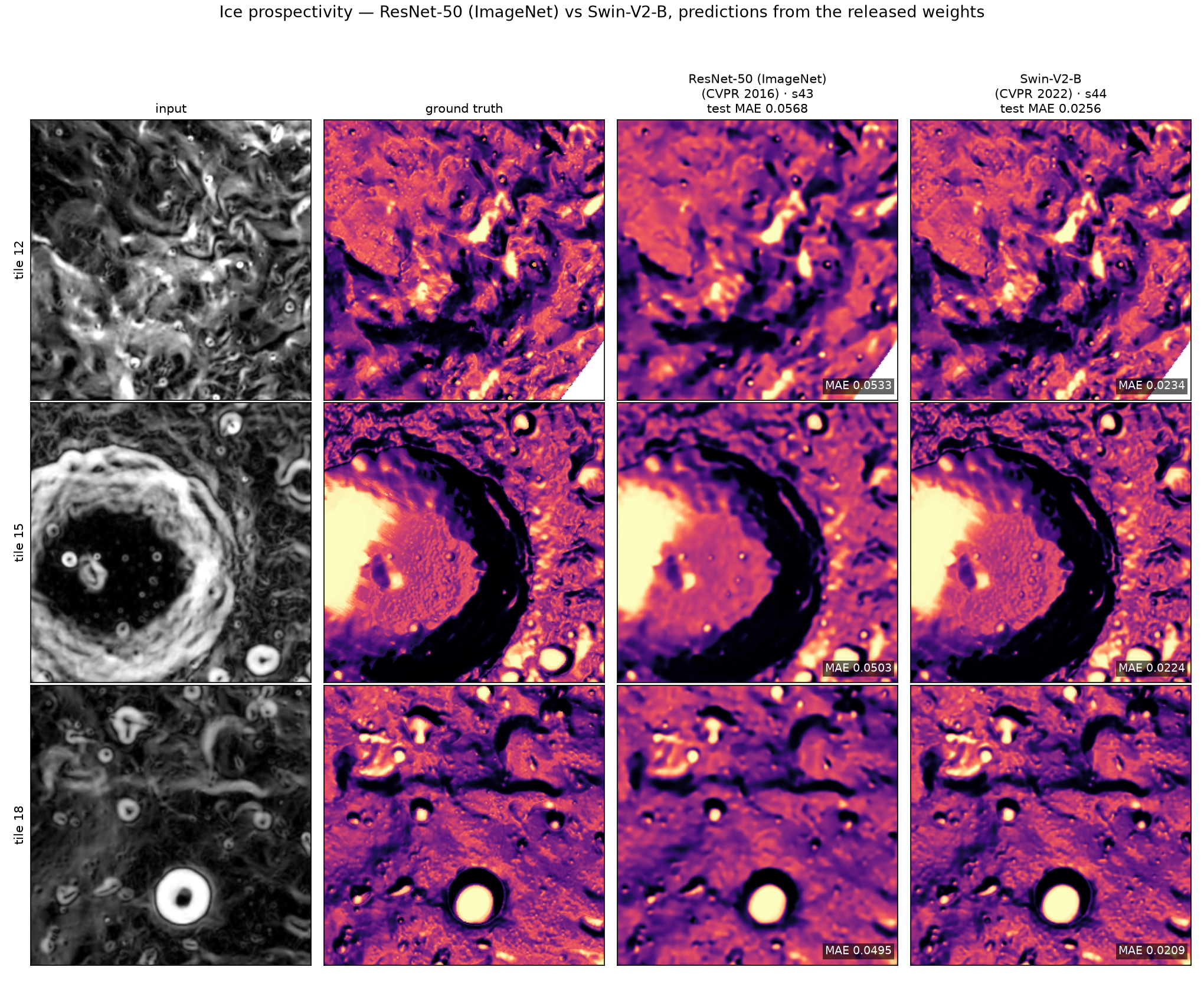}
    \caption{Qualitative ice-prospectivity regression results on three polar test patches. Columns show a representative input layer, the target prospectivity map, and predictions from ResNet-50 (ImageNet) and SwinV2-B, using the released weights; per-panel insets report the per-patch MAE.}
    \label{fig:ice-prosp-qual}
\end{figure}

The ice-prospectivity benchmark is framed as per-pixel regression, with eight evidential layers of the polar stack (slope, curvature, aspect encoded as a sine--cosine pair, maximum surface temperature, ice stability depth, PSRs, distance to PSRs, and PSR density) form the input, and the prospectivity map is the target. Each modality is normalized independently and backbone features are combined through a multimodal fusion decoder. Models are trained for 100 epochs and selected on validation root mean squared error (RMSE). Table~\ref{tab:downstream-ice-prosp} reports test-set RMSE, mean absolute error (MAE), and coefficient of determination ($R^{2}$), averaged over five random seeds. Both models recover the prospectivity field with high fidelity ($R^{2} \geq 0.92$). Figure~\ref{fig:ice-prosp-qual} shows qualitative predictions on three polar test patches.

\begin{table}[H]
\centering
\caption{Polar ice prospectivity regression on the eight-modality polar stack (aspect, slope, ice stability depth, maximum temperature, PSRs, PSR density, distance to PSRs, curvature). Test-set metrics against the prospectivity map in \cite{coyan2025prospectivity}: for RMSE and MAE lower-is-better, and for $R^{2}$ higher-is-better. Values are mean $\pm$ standard deviation over five random seeds.}
\label{tab:downstream-ice-prosp}
\begin{tabular}{lccc}
\toprule
Model & RMSE $\downarrow$ & MAE $\downarrow$ & $R^{2}$ $\uparrow$ \\
\midrule
ResNet-50 (ImageNet)   & 0.0788 $\pm$ 0.0004 & 0.0568 $\pm$ 0.0004 & 0.9162 $\pm$ 0.0009 \\
SwinV2-B (ImageNet)    & 0.0377 $\pm$ 0.0004 & 0.0256 $\pm$ 0.0003 & 0.9808 $\pm$ 0.0004 \\
\bottomrule
\end{tabular}
\end{table}

\section{Usage Notes and Known Limitations} 

\hspace*{2em}This section summarizes practical considerations that affect interpretation and downstream use of \dataset: how the released pre-training corpus and benchmarks should be consumed, the harmonization limits inherited from multi-instrument sources, and the geodetic accuracy caveats that matter for any analysis requiring high-precision localization.

\subsection*{Working with the Released Corpus}

The following notes concern the released pre-training and benchmark artifacts themselves (Sections~\ref{sec:construction} and \ref{sec:data-records}):

\begin{itemize}[nosep,leftmargin=1.2em]
  \item \textbf{Split discipline.} The train and validation splits are intended for pre-training. The test split is reserved for evaluation and downstream fine-tuning (Section~\ref{ia:train-test-val-split}), and benchmark tiles drawn from the corpus come from this held-out test split. Pre-training on test-split tiles invalidates benchmark comparisons, and any custom split should preserve the zone-wise partition to avoid spatial leakage.
  \item \textbf{Illumination geometry is a confounder.} Apparent surface contrast on the Moon is dominated by incidence angle and solar azimuth, so models can learn lighting rather than geology. The per-tile viewing-geometry columns in the Parquet catalogs (incidence, emission, and phase angles and sub-solar geometry) should be used to stratify training data, audit evaluation results, or condition models.
  \item \textbf{NaN semantics.} NaN encodes a genuine data gap and is preserved through value-range clipping. Layers exceeding the 90\% NaN gate are recorded as missing rather than filled. Users should consult the per-modality \texttt{FRACTION\_NULL} catalog columns and handle gaps explicitly rather than imputing silently.
  \item \textbf{Polar-only modalities are sparse.} Layers restricted to the poles (PSRs, average illumination, 1064 nm albedo, ice stability depth, and hydrogen abundance) exist for fewer than 1\% of tiles in either track (Table~\ref{tab:modality-coverage-and-resolution}). They should be treated as conditionally available inputs, not as globally present channels.
  \item \textbf{Tile redundancy.} Because tiles are anchored to individual EDR canvases, a ground patch covered by $m$ admitted EDRs is materialized $m$ times (Section~\ref{sec:construction}). Spatial statistics, per-region evaluations, and dataset subsampling should therefore deduplicate by ground footprint rather than by tile count.
\end{itemize}

\subsection*{Data Harmonization Challenges}

Although many inputs originate from a single mission family (primarily LRO), “single mission” does not imply “single format.” \dataset integrates products that differ in spatial scale, coverage patterns, file conventions, and noise/artifact structure; the preprocessing of Section~\ref{subsec:preprocessing} standardizes projections and grids, but the following differences are inherent to the sources:

\begin{itemize}[nosep,leftmargin=1.2em]
  \item \textbf{Heterogeneous spatial resolution.} Inputs span orders of magnitude in pixel scale, from meter-scale targeted imaging through tens to hundreds of meters per pixel global products to coarser static context layers. Many ML pipelines implicitly assume comparable footprints and stable sampling density; multimodal fusion models should handle scale mismatch explicitly (e.g., tiling, pyramids, learned up/down sampling, or modality-specific encoders).
  \item \textbf{Heterogeneous temporal sampling and coverage.} Some layers are global static mosaics, while others aggregate time-dependent sampling into static products. Layer-to-layer differences should not be interpreted as change over time unless the product is explicitly time-resolved.
  \item \textbf{Product lineage and conventions.} Lunar products are distributed under instrument-specific pipelines and record types (e.g., calibrated versus reduced/derived products) with differing metadata conventions. When combining layers, users should verify that inputs are comparable in meaning (e.g., reflectance versus relative brightness; photometrically normalized versus illumination-selected mosaics).
  \item \textbf{Label scarcity and class imbalance.} Many lunar targets lack dense, globally uniform annotations, which limits purely supervised approaches and can bias benchmarks toward well-labeled regions or feature types. Several benchmark tasks are inherently imbalanced (rare features, restricted geographic distributions, or limited illumination regimes), so evaluation should use metrics robust to imbalance.
  \item \textbf{Data volume and instrument-specific artifacts.} The combined archive is large enough to introduce practical bottlenecks (I/O, preprocessing time, storage), and instrument- or product-specific artifacts (illumination variability, seam artifacts, radar layover-like effects, thermal model assumptions) can drive spurious correlations unless models are trained with appropriate controls and quality filters.
\end{itemize}

\subsection*{Location Precision and Geodetic Accuracy}

High-resolution imagery does not guarantee high-confidence coordinates. Practical offsets and accuracy dilution can arise from the choice of reference frames, ephemerides, control status (controlled versus uncontrolled products), and topography and viewing geometry:

\begin{itemize}[nosep,leftmargin=1.2em]
  \item \textbf{Limited absolute ground control.} The Moon lacks a dense, globally surveyed network of ground control points; absolute anchors are limited to a small number of retroreflector sites, which constrains global absolute accuracy.
  \item \textbf{Reference frames and ephemerides.} Products can be tied to different reference frame realizations (e.g., DE421-era products versus newer ephemerides), and mixing lineages can introduce systematic offsets and dilute overall positional accuracy if not handled consistently.
  \item \textbf{Controlled versus uncontrolled products.} Uncontrolled products can carry mission-phase-dependent offsets driven by spacecraft trajectory uncertainty. Controlled mosaics and DTMs achieve substantially better internal consistency through co-registration to reference datasets and bundle adjustment, but residual offsets at the tens-of-meters level can remain depending on product lineage.
  \item \textbf{Topography and viewing geometry.} Off-nadir viewing and steep relief can introduce pixel-scale parallax unless imagery is orthorectified to an appropriate-resolution DTM, and misregistration risk increases when high-resolution imagery is paired with lower-resolution terrain models.
\end{itemize}

For analyses that depend on precise locations or fine-scale co-registration, we recommend the following practices, adapted from \cite{wagner2024crater}:

\begin{itemize}[nosep,leftmargin=1.2em]
  \item \textbf{Prefer the most controlled products available.} Use orthorectified and controlled imagery and DTMs where available, and treat uncontrolled images as lower confidence for absolute position.
  \item \textbf{Report coordinates at realistic precision.} Avoid implying sub-meter accuracy through excessive decimals. Report coordinate precision consistent with product control status (e.g., fewer decimals for uncontrolled products).
  \item \textbf{Include reproducibility metadata.} Provide annotated context imagery for coordinate picks, and when using unprojected images, report pixel line/sample (or equivalent) to enable exact reproduction.
  \item \textbf{Use appropriate ephemerides and terrain models.} When high precision is required, use refined spacecraft ephemerides when available and orthorectify to the best available local DTM rather than a coarse global terrain model.
  \item \textbf{Cross-check across observations.} Where controlled data are unavailable, estimate locations from multiple suitable observations and compare results to identify outliers and systematic offsets.
\end{itemize}

Taken together, these considerations motivate \dataset’s emphasis on standardized projections and metadata conventions, clear separation of global versus polar product variants, and documentation of processing lineage so that benchmark results reflect geologic and physical signal rather than registration artifacts.


\section*{Competing Interests}
The authors declare no competing interests.

\section*{Acknowledgments}
This work is supported by NASA Grant 80MSFC22M004. The Authors acknowledge the National Artificial Intelligence Research Resource (NAIRR) Pilot and NVIDIA for providing support under grant no. NAIRR240178. We also want to thank Ethan Anderson, Anujan Ganeshalingam, Onkar Hunjan, and Gwilym Newton for their assistance in hand circling craters. VV and ZM are supported by NASA under CRESST II Cooperative Agreement (80GSFC24M0006).

\section*{Author Contributions}

\textbf{H. Patil}: Methodology, Software, Investigation, Formal analysis, Data curation, Visualization, Writing - Original Draft

\textbf{G. Nyirjesy}: Methodology, Investigation, Formal analysis, Data curation, Visualization, Writing - Original Draft 

\textbf{R. A. Slank}: Methodology, Validation, Formal Analysis, Data Curation, Writing [crater dataset]

\textbf{V. Gaur}: Methodology, Software, Investigation, Formal analysis, Data curation, ML Benchmarks, Visualization, Writing - Original Draft

\textbf{D. Szwarcman}: Data curation, Visualization, Writing - Review

\textbf{P. Fraccaro}: Data curation, Visualization, Writing - Review

\textbf{N. Dionelis}: Data curation, Visualization, Writing - Review

\textbf{M. K. Barker}: Conceptualization, Methodology, Software, Validation, Formal analysis, Data curation, Writing - Review \& Editing, Supervision, Science - Project administration

\textbf{A. M. Annex}: Conceptualization, Methodology, Software, Validation, Formal analysis, Investigation, Data Curation, Writing - Review \& Editing

\textbf{V. Viswanathan}: Methodology, Validation, Formal analysis, Data Curation, Writing - Review \& Editing

\textbf{Z. Morse}: Methodology, Validation, Formal analysis, Data Curation, Writing - Review \& Editing

\textbf{E. I. Schaefer}: Methodology, Software, Validation, Formal analysis, Data Curation, Writing - Review \& Editing

\textbf{H. Debary}: Data curation, Visualization, ML benchmarks, Writing - Review \& Editing

\textbf{A. Kumar}: Data curation, Writing - Review

\textbf{R. Lal}: ML benchmarks, Writing - Review  \& Editing

\textbf{G. Dawson}: ML benchmarks, Writing - Review  \& Editing

\textbf{C. Watson}: Writing - Review  \& Editing

\textbf{R. I. Dawson-Rigas}: Writing - Review  \& Editing, Project administration

\textbf{M. Maskey}: ML benchmarks, Writing - Review  \& Editing

\textbf{J. B. Moreno}: Writing - Review  \& Editing, Project administration

\textbf{R. Ramachandran}: Writing - Review  \& Editing, Project administration

\textbf{S. Roy}: Conceptualization, Methodology, Software, Validation, Formal analysis, Data curation, Writing - Review, Supervision, FM Project administration



\section*{Acknowledgements}
This work was supported by the National Aeronautics and Space Administration under Award No. 80MSFC25M0084. The Authors acknowledge the National Artificial Intelligence Research Resource (NAIRR) Pilot and Voltage Park for providing support under grant no. NAIRR250201. 
\bibliography{references}

\begin{thebibliography}{10}
\urlstyle{rm}
\expandafter\ifx\csname url\endcsname\relax
  \def\url#1{\texttt{#1}}\fi
\expandafter\ifx\csname urlprefix\endcsname\relax\def\urlprefix{URL }\fi
\expandafter\ifx\csname doiprefix\endcsname\relax\def\doiprefix{DOI: }\fi
\providecommand{\bibinfo}[2]{#2}
\providecommand{\eprint}[2][]{\url{#2}}

\bibitem{robinson2010lunar}
\bibinfo{author}{Robinson, M.~S.} \emph{et~al.}
\newblock \bibinfo{journal}{\bibinfo{title}{Lunar reconnaissance orbiter camera (lroc) instrument overview}}.
\newblock {\emph{\JournalTitle{Space science reviews}}} \textbf{\bibinfo{volume}{150}}, \bibinfo{pages}{81--124} (\bibinfo{year}{2010}).

\bibitem{smith2010lunar}
\bibinfo{author}{Smith, D.~E.} \emph{et~al.}
\newblock \bibinfo{journal}{\bibinfo{title}{The lunar orbiter laser altimeter investigation on the lunar reconnaissance orbiter mission}}.
\newblock {\emph{\JournalTitle{Space science reviews}}} \textbf{\bibinfo{volume}{150}}, \bibinfo{pages}{209--241} (\bibinfo{year}{2010}).

\bibitem{paige2010lunar}
\bibinfo{author}{Paige, D.} \emph{et~al.}
\newblock \bibinfo{journal}{\bibinfo{title}{The lunar reconnaissance orbiter diviner lunar radiometer experiment}}.
\newblock {\emph{\JournalTitle{Space Science Reviews}}} \textbf{\bibinfo{volume}{150}}, \bibinfo{pages}{125--160} (\bibinfo{year}{2010}).

\bibitem{nozette2010lunar}
\bibinfo{author}{Nozette, S.} \emph{et~al.}
\newblock \bibinfo{journal}{\bibinfo{title}{The lunar reconnaissance orbiter miniature radio frequency (mini-rf) technology demonstration}}.
\newblock {\emph{\JournalTitle{Space Science Reviews}}} \textbf{\bibinfo{volume}{150}}, \bibinfo{pages}{285--302} (\bibinfo{year}{2010}).

\bibitem{lemelin2015lunar}
\bibinfo{author}{Lemelin, M.}, \bibinfo{author}{Lucey, P.~G.}, \bibinfo{author}{Song, E.} \& \bibinfo{author}{Taylor, G.~J.}
\newblock \bibinfo{journal}{\bibinfo{title}{Lunar central peak mineralogy and iron content using the kaguya multiband imager: Reassessment of the compositional structure of the lunar crust}}.
\newblock {\emph{\JournalTitle{Journal of Geophysical Research: Planets}}} \textbf{\bibinfo{volume}{120}}, \bibinfo{pages}{869--887} (\bibinfo{year}{2015}).

\bibitem{lemelin2016global}
\bibinfo{author}{Lemelin, M.}, \bibinfo{author}{Lucey, P.}, \bibinfo{author}{Gaddis, L.}, \bibinfo{author}{Hare, T.} \& \bibinfo{author}{Ohtake, M.}
\newblock \bibinfo{title}{Global map products from the kaguya multiband imager at 512 ppd: Minerals, feo, and omat}.
\newblock In \emph{\bibinfo{booktitle}{47th annual lunar and planetary science conference}}, \bibinfo{number}{1903}, \bibinfo{pages}{2994} (\bibinfo{year}{2016}).

\bibitem{lemelin2019compositions}
\bibinfo{author}{Lemelin, M.} \emph{et~al.}
\newblock \bibinfo{journal}{\bibinfo{title}{The compositions of the lunar crust and upper mantle: Spectral analysis of the inner rings of lunar impact basins}}.
\newblock {\emph{\JournalTitle{Planetary and Space Science}}} \textbf{\bibinfo{volume}{165}}, \bibinfo{pages}{230--243} (\bibinfo{year}{2019}).

\bibitem{barker2016new}
\bibinfo{author}{Barker, M.} \emph{et~al.}
\newblock \bibinfo{journal}{\bibinfo{title}{A new lunar digital elevation model from the lunar orbiter laser altimeter and selene terrain camera}}.
\newblock {\emph{\JournalTitle{Icarus}}} \textbf{\bibinfo{volume}{273}}, \bibinfo{pages}{346--355} (\bibinfo{year}{2016}).

\bibitem{zuber2013gravity}
\bibinfo{author}{Zuber, M.~T.} \emph{et~al.}
\newblock \bibinfo{journal}{\bibinfo{title}{Gravity recovery and interior laboratory (grail): Mapping the lunar interior from crust to core}}.
\newblock {\emph{\JournalTitle{Space Science Reviews}}} \textbf{\bibinfo{volume}{178}}, \bibinfo{pages}{3--24} (\bibinfo{year}{2013}).

\bibitem{konopliv2014high}
\bibinfo{author}{Konopliv, A.~S.} \emph{et~al.}
\newblock \bibinfo{journal}{\bibinfo{title}{High-resolution lunar gravity fields from the grail primary and extended missions}}.
\newblock {\emph{\JournalTitle{Geophysical Research Letters}}} \textbf{\bibinfo{volume}{41}}, \bibinfo{pages}{1452--1458} (\bibinfo{year}{2014}).

\bibitem{feldman1999lunar}
\bibinfo{author}{Feldman, W.} \emph{et~al.}
\newblock \bibinfo{journal}{\bibinfo{title}{The lunar prospector gamma-ray and neutron spectrometers}}.
\newblock {\emph{\JournalTitle{Nuclear Instruments and Methods in Physics Research Section A: Accelerators, Spectrometers, Detectors and Associated Equipment}}} \textbf{\bibinfo{volume}{422}}, \bibinfo{pages}{562--566} (\bibinfo{year}{1999}).

\bibitem{feldman2001evidence}
\bibinfo{author}{Feldman, W.~C.} \emph{et~al.}
\newblock \bibinfo{journal}{\bibinfo{title}{Evidence for water ice near the lunar poles}}.
\newblock {\emph{\JournalTitle{Journal of Geophysical Research: Planets}}} \textbf{\bibinfo{volume}{106}}, \bibinfo{pages}{23231--23251} (\bibinfo{year}{2001}).

\bibitem{cong2022satmae}
\bibinfo{author}{Cong, Y.} \emph{et~al.}
\newblock \bibinfo{journal}{\bibinfo{title}{Satmae: Pre-training transformers for temporal and multi-spectral satellite imagery}}.
\newblock {\emph{\JournalTitle{Advances in Neural Information Processing Systems}}} \textbf{\bibinfo{volume}{35}}, \bibinfo{pages}{197--211} (\bibinfo{year}{2022}).

\bibitem{stewart2023ssl4eo}
\bibinfo{author}{Stewart, A.} \emph{et~al.}
\newblock \bibinfo{journal}{\bibinfo{title}{Ssl4eo-l: Datasets and foundation models for landsat imagery}}.
\newblock {\emph{\JournalTitle{Advances in Neural Information Processing Systems}}} \textbf{\bibinfo{volume}{36}}, \bibinfo{pages}{59787--59807} (\bibinfo{year}{2023}).

\bibitem{fuller2023croma}
\bibinfo{author}{Fuller, A.}, \bibinfo{author}{Millard, K.} \& \bibinfo{author}{Green, J.}
\newblock \bibinfo{journal}{\bibinfo{title}{Croma: Remote sensing representations with contrastive radar-optical masked autoencoders}}.
\newblock {\emph{\JournalTitle{Advances in Neural Information Processing Systems}}} \textbf{\bibinfo{volume}{36}}, \bibinfo{pages}{5506--5538} (\bibinfo{year}{2023}).

\bibitem{reed2023scale}
\bibinfo{author}{Reed, C.~J.} \emph{et~al.}
\newblock \bibinfo{title}{Scale-mae: A scale-aware masked autoencoder for multiscale geospatial representation learning}.
\newblock In \emph{\bibinfo{booktitle}{Proceedings of the IEEE/CVF International Conference on Computer Vision}}, \bibinfo{pages}{4088--4099} (\bibinfo{year}{2023}).

\bibitem{roy2025suryabench}
\bibinfo{author}{Roy, S.} \emph{et~al.}
\newblock \bibinfo{journal}{\bibinfo{title}{Suryabench: Benchmark dataset for advancing machine learning in heliophysics and space weather prediction}}.
\newblock {\emph{\JournalTitle{arXiv preprint arXiv:2508.14107}}}  (\bibinfo{year}{2025}).

\bibitem{prasad2026moonstonemultimodalfoundationmodel}
\bibinfo{author}{Prasad, A.} \& \bibinfo{author}{Mazumder, S.}
\newblock \bibinfo{title}{Moonstone: A multimodal foundation model and benchmark for lunar remote sensing} (\bibinfo{year}{2026}).
\newblock \eprint{2607.03644}.

\bibitem{chin2007lunar}
\bibinfo{author}{Chin, G.} \emph{et~al.}
\newblock \bibinfo{journal}{\bibinfo{title}{Lunar reconnaissance orbiter overview: The instrument suite and mission}}.
\newblock {\emph{\JournalTitle{Space Science Reviews}}} \textbf{\bibinfo{volume}{129}}, \bibinfo{pages}{391--419} (\bibinfo{year}{2007}).

\bibitem{vondrak2010lunar}
\bibinfo{author}{Vondrak, R.}, \bibinfo{author}{Keller, J.}, \bibinfo{author}{Chin, G.} \& \bibinfo{author}{Garvin, J.}
\newblock \bibinfo{journal}{\bibinfo{title}{Lunar reconnaissance orbiter (lro): Observations for lunar exploration and science}}.
\newblock {\emph{\JournalTitle{Space Science Reviews}}} \textbf{\bibinfo{volume}{150}}, \bibinfo{pages}{7--22} (\bibinfo{year}{2010}).

\bibitem{raney2007hybrid}
\bibinfo{author}{Raney, R.~K.}
\newblock \bibinfo{journal}{\bibinfo{title}{Hybrid-polarity sar architecture}}.
\newblock {\emph{\JournalTitle{IEEE Transactions on Geoscience and Remote Sensing}}} \textbf{\bibinfo{volume}{45}}, \bibinfo{pages}{3397--3404} (\bibinfo{year}{2007}).

\bibitem{kato2010kaguya}
\bibinfo{author}{Kato, M.}, \bibinfo{author}{Sasaki, S.}, \bibinfo{author}{Takizawa, Y.} \& \bibinfo{author}{the Kaguya~project team}.
\newblock \bibinfo{journal}{\bibinfo{title}{The kaguya mission overview}}.
\newblock {\emph{\JournalTitle{Space Science Reviews}}} \textbf{\bibinfo{volume}{154}}, \bibinfo{pages}{3--19} (\bibinfo{year}{2010}).

\bibitem{ohtake2008performance}
\bibinfo{author}{Ohtake, M.} \emph{et~al.}
\newblock \bibinfo{journal}{\bibinfo{title}{Performance and scientific objectives of the selene (kaguya) multiband imager}}.
\newblock {\emph{\JournalTitle{Earth, planets and space}}} \textbf{\bibinfo{volume}{60}}, \bibinfo{pages}{257--264} (\bibinfo{year}{2008}).

\bibitem{haruyama2008global}
\bibinfo{author}{Haruyama, J.} \emph{et~al.}
\newblock \bibinfo{journal}{\bibinfo{title}{Global lunar-surface mapping experiment using the lunar imager/spectrometer on selene}}.
\newblock {\emph{\JournalTitle{Earth, planets and space}}} \textbf{\bibinfo{volume}{60}}, \bibinfo{pages}{243--255} (\bibinfo{year}{2008}).

\bibitem{yamamoto2014calibration}
\bibinfo{author}{Yamamoto, S.} \emph{et~al.}
\newblock \bibinfo{journal}{\bibinfo{title}{Calibration of nir 2 of spectral profiler onboard kaguya/selene}}.
\newblock {\emph{\JournalTitle{IEEE Transactions on Geoscience and Remote Sensing}}} \textbf{\bibinfo{volume}{52}}, \bibinfo{pages}{6882--6898} (\bibinfo{year}{2014}).

\bibitem{binder1998lunar}
\bibinfo{author}{Binder, A.~B.}
\newblock \bibinfo{journal}{\bibinfo{title}{Lunar prospector: Overview}}.
\newblock {\emph{\JournalTitle{Science}}} \textbf{\bibinfo{volume}{281}}, \bibinfo{pages}{1475--1476} (\bibinfo{year}{1998}).

\bibitem{Lawrence2022}
\bibinfo{author}{Lawrence, D.~J.}, \bibinfo{author}{Peplowski, P.~N.}, \bibinfo{author}{Wilson, J.~T.} \& \bibinfo{author}{Elphic, R.~C.}
\newblock \bibinfo{journal}{\bibinfo{title}{Global hydrogen abundances on the lunar surface}}.
\newblock {\emph{\JournalTitle{Journal of Geophysical Research: Planets}}} \textbf{\bibinfo{volume}{127}}, \bibinfo{pages}{e2022JE007197}, \url{https://doi.org/10.1029/2022JE007197} (\bibinfo{year}{2022}).
\newblock \bibinfo{note}{E2022JE007197 2022JE007197}, \eprint{https://agupubs.onlinelibrary.wiley.com/doi/pdf/10.1029/2022JE007197}.

\bibitem{beyer2018ames}
\bibinfo{author}{Beyer, R.~A.}, \bibinfo{author}{Alexandrov, O.} \& \bibinfo{author}{McMichael, S.}
\newblock \bibinfo{journal}{\bibinfo{title}{The ames stereo pipeline: Nasa's open source software for deriving and processing terrain data}}.
\newblock {\emph{\JournalTitle{Earth and Space Science}}} \textbf{\bibinfo{volume}{5}}, \bibinfo{pages}{537--548} (\bibinfo{year}{2018}).

\bibitem{henriksen2017extracting}
\bibinfo{author}{Henriksen, M.} \emph{et~al.}
\newblock \bibinfo{journal}{\bibinfo{title}{Extracting accurate and precise topography from lroc narrow angle camera stereo observations}}.
\newblock {\emph{\JournalTitle{Icarus}}} \textbf{\bibinfo{volume}{283}}, \bibinfo{pages}{122--137} (\bibinfo{year}{2017}).

\bibitem{kirk2008ultrahigh}
\bibinfo{author}{Kirk, R.~L.} \emph{et~al.}
\newblock \bibinfo{journal}{\bibinfo{title}{Ultrahigh resolution topographic mapping of mars with mro hirise stereo images: Meter-scale slopes of candidate phoenix landing sites}}.
\newblock {\emph{\JournalTitle{Journal of Geophysical Research: Planets}}} \textbf{\bibinfo{volume}{113}} (\bibinfo{year}{2008}).

\bibitem{speyerer2011lunar}
\bibinfo{author}{Speyerer, E.}, \bibinfo{author}{Robinson, M.}, \bibinfo{author}{Denevi, B.}, \bibinfo{author}{Team, L.~S.} \emph{et~al.}
\newblock \bibinfo{title}{Lunar reconnaissance orbiter camera global morphological map of the moon}.
\newblock In \emph{\bibinfo{booktitle}{42nd annual lunar and planetary science conference}}, \bibinfo{number}{1608}, \bibinfo{pages}{2387} (\bibinfo{year}{2011}).

\bibitem{robinson2012exploring}
\bibinfo{author}{Robinson, M.} \emph{et~al.}
\newblock \bibinfo{journal}{\bibinfo{title}{Exploring the moon with the lunar reconnaissance orbiter camera}}.
\newblock {\emph{\JournalTitle{The International Archives of the Photogrammetry, Remote Sensing and Spatial Information Sciences}}} \textbf{\bibinfo{volume}{39}}, \bibinfo{pages}{501--504} (\bibinfo{year}{2012}).

\bibitem{wagner2015new}
\bibinfo{author}{Wagner, R.}, \bibinfo{author}{Speyerer, E.}, \bibinfo{author}{Robinson, M.}, \bibinfo{author}{team, L.} \emph{et~al.}
\newblock \bibinfo{title}{New mosaicked data products from the lroc team}.
\newblock In \emph{\bibinfo{booktitle}{46th annual lunar and planetary science conference}}, \bibinfo{number}{1832}, \bibinfo{pages}{1473} (\bibinfo{year}{2015}).

\bibitem{speyerer2013persistently}
\bibinfo{author}{Speyerer, E.~J.} \& \bibinfo{author}{Robinson, M.~S.}
\newblock \bibinfo{journal}{\bibinfo{title}{Persistently illuminated regions at the lunar poles: Ideal sites for future exploration}}.
\newblock {\emph{\JournalTitle{Icarus}}} \textbf{\bibinfo{volume}{222}}, \bibinfo{pages}{122--136} (\bibinfo{year}{2013}).

\bibitem{sato2017lunar}
\bibinfo{author}{Sato, H.} \emph{et~al.}
\newblock \bibinfo{journal}{\bibinfo{title}{Lunar mare tio2 abundances estimated from uv/vis reflectance}}.
\newblock {\emph{\JournalTitle{Icarus}}} \textbf{\bibinfo{volume}{296}}, \bibinfo{pages}{216--238} (\bibinfo{year}{2017}).

\bibitem{boyd2012lunar}
\bibinfo{author}{Boyd, A.}, \bibinfo{author}{Robinson, M.} \& \bibinfo{author}{Sato, H.}
\newblock \bibinfo{title}{Lunar reconnaissance orbiter wide angle camera photometry: An empirical solution}.
\newblock In \emph{\bibinfo{booktitle}{43rd Annual Lunar and Planetary Science Conference}}, \bibinfo{number}{1659}, \bibinfo{pages}{2795} (\bibinfo{year}{2012}).

\bibitem{sato2014resolved}
\bibinfo{author}{Sato, H.}, \bibinfo{author}{Robinson, M.}, \bibinfo{author}{Hapke, B.}, \bibinfo{author}{Denevi, B.} \& \bibinfo{author}{Boyd, A.}
\newblock \bibinfo{journal}{\bibinfo{title}{Resolved hapke parameter maps of the moon}}.
\newblock {\emph{\JournalTitle{Journal of Geophysical Research: Planets}}} \textbf{\bibinfo{volume}{119}}, \bibinfo{pages}{1775--1805} (\bibinfo{year}{2014}).

\bibitem{Trang2019}
\bibinfo{author}{Trang, D.} \& \bibinfo{author}{Lucey, P.~G.}
\newblock \bibinfo{journal}{\bibinfo{title}{Improved space weathering maps of the lunar surface through radiative transfer modeling of kaguya multiband imager data}}.
\newblock {\emph{\JournalTitle{Icarus}}} \textbf{\bibinfo{volume}{321}}, \bibinfo{pages}{307--323}, \url{https://doi.org/10.1016/j.icarus.2018.11.014} (\bibinfo{year}{2019}).

\bibitem{barker2023}
\bibinfo{author}{{Barker}, M.~K.} \emph{et~al.}
\newblock \bibinfo{journal}{\bibinfo{title}{{A New View of the Lunar South Pole from the Lunar Orbiter Laser Altimeter (LOLA)}}}.
\newblock {\emph{\JournalTitle{\psj}}} \textbf{\bibinfo{volume}{4}}, \bibinfo{pages}{183}, \url{10.3847/PSJ/acf3e1} (\bibinfo{year}{2023}).

\bibitem{barker2025large}
\bibinfo{author}{Barker, M.~K.} \emph{et~al.}
\newblock \bibinfo{journal}{\bibinfo{title}{Large-scale roughness properties of the lunar north and south polar regions as measured by the lunar orbiter laser altimeter (lola)}}.
\newblock {\emph{\JournalTitle{The Planetary Science Journal}}} \textbf{\bibinfo{volume}{6}}, \bibinfo{pages}{83} (\bibinfo{year}{2025}).

\bibitem{kreslavsky2013lunar}
\bibinfo{author}{Kreslavsky, M.~A.} \emph{et~al.}
\newblock \bibinfo{journal}{\bibinfo{title}{Lunar topographic roughness maps from lunar orbiter laser altimeter (lola) data: Scale dependence and correlation with geologic features and units}}.
\newblock {\emph{\JournalTitle{Icarus}}} \textbf{\bibinfo{volume}{226}}, \bibinfo{pages}{52--66} (\bibinfo{year}{2013}).

\bibitem{neumann2015}
\bibinfo{author}{{Neumann}, G.~A.}, \bibinfo{author}{{Glaeser}, P.~A.}, \bibinfo{author}{{Hiesinger}, H.}, \bibinfo{author}{{Zuber}, M.~T.} \& \bibinfo{author}{{Smith}, D.~E.}
\newblock \bibinfo{title}{{Copernican-Age Craters and LOLA Decameter-Scale Roughness}}.
\newblock In \emph{\bibinfo{booktitle}{46th Annual Lunar and Planetary Science Conference}}, Lunar and Planetary Science Conference, \bibinfo{pages}{2218} (\bibinfo{year}{2015}).

\bibitem{mazarico2011illumination}
\bibinfo{author}{Mazarico, E.}, \bibinfo{author}{Neumann, G.}, \bibinfo{author}{Smith, D.}, \bibinfo{author}{Zuber, M.} \& \bibinfo{author}{Torrence, M.}
\newblock \bibinfo{journal}{\bibinfo{title}{Illumination conditions of the lunar polar regions using lola topography}}.
\newblock {\emph{\JournalTitle{Icarus}}} \textbf{\bibinfo{volume}{211}}, \bibinfo{pages}{1066--1081} (\bibinfo{year}{2011}).

\bibitem{Lemelin2016}
\bibinfo{author}{{Lemelin}, M.} \emph{et~al.}
\newblock \bibinfo{journal}{\bibinfo{title}{{Improved calibration of reflectance data from the LRO Lunar Orbiter Laser Altimeter (LOLA) and implications for space weathering}}}.
\newblock {\emph{\JournalTitle{\icarus}}} \textbf{\bibinfo{volume}{273}}, \bibinfo{pages}{315--328}, \url{10.1016/j.icarus.2016.02.006} (\bibinfo{year}{2016}).

\bibitem{fassett2024}
\bibinfo{author}{{Fassett}, C.~I.} \emph{et~al.}
\newblock \bibinfo{journal}{\bibinfo{title}{{Improved Orthorectification and Empirical Reduction of Topographic Effects in Monostatic Mini-RF S-band Observations of the Moon}}}.
\newblock {\emph{\JournalTitle{\psj}}} \textbf{\bibinfo{volume}{5}}, \bibinfo{pages}{4}, \url{10.3847/PSJ/ad0a61} (\bibinfo{year}{2024}).

\bibitem{raney2010lunar}
\bibinfo{author}{Raney, R.~K.} \emph{et~al.}
\newblock \bibinfo{journal}{\bibinfo{title}{The lunar mini-rf radars: Hybrid polarimetric architecture and initial results}}.
\newblock {\emph{\JournalTitle{Proceedings of the IEEE}}} \textbf{\bibinfo{volume}{99}}, \bibinfo{pages}{808--823} (\bibinfo{year}{2010}).

\bibitem{spudis2013evidence}
\bibinfo{author}{Spudis, P.} \emph{et~al.}
\newblock \bibinfo{journal}{\bibinfo{title}{Evidence for water ice on the moon: Results for anomalous polar craters from the lro mini-rf imaging radar}}.
\newblock {\emph{\JournalTitle{Journal of Geophysical Research: Planets}}} \textbf{\bibinfo{volume}{118}}, \bibinfo{pages}{2016--2029} (\bibinfo{year}{2013}).

\bibitem{neish2011surficial}
\bibinfo{author}{Neish, C.} \emph{et~al.}
\newblock \bibinfo{journal}{\bibinfo{title}{The surficial nature of lunar swirls as revealed by the mini-rf instrument}}.
\newblock {\emph{\JournalTitle{Icarus}}} \textbf{\bibinfo{volume}{215}}, \bibinfo{pages}{186--196} (\bibinfo{year}{2011}).

\bibitem{campbell1997regolith}
\bibinfo{author}{Campbell, B.~A.}, \bibinfo{author}{Hawke, B.~R.} \& \bibinfo{author}{Thompson, T.~W.}
\newblock \bibinfo{journal}{\bibinfo{title}{Regolith composition and structure in the lunar maria: Results of long-wavelength radar studies}}.
\newblock {\emph{\JournalTitle{Journal of Geophysical Research: Planets}}} \textbf{\bibinfo{volume}{102}}, \bibinfo{pages}{19307--19320} (\bibinfo{year}{1997}).

\bibitem{fa2013circular}
\bibinfo{author}{Fa, W.} \& \bibinfo{author}{Cai, Y.}
\newblock \bibinfo{journal}{\bibinfo{title}{Circular polarization ratio characteristics of impact craters from mini-rf observations and implications for ice detection at the polar regions of the moon}}.
\newblock {\emph{\JournalTitle{Journal of Geophysical Research: Planets}}} \textbf{\bibinfo{volume}{118}}, \bibinfo{pages}{1582--1608} (\bibinfo{year}{2013}).

\bibitem{Fa2018}
\bibinfo{author}{Fa, W.} \& \bibinfo{author}{Eke, V.~R.}
\newblock \bibinfo{journal}{\bibinfo{title}{Unravelling the mystery of lunar anomalous craters using radar and infrared observations}}.
\newblock {\emph{\JournalTitle{Journal of Geophysical Research: Planets}}} \textbf{\bibinfo{volume}{123}}, \bibinfo{pages}{2119--2137}, \url{https://doi.org/10.1029/2018JE005668} (\bibinfo{year}{2018}).
\newblock \eprint{https://agupubs.onlinelibrary.wiley.com/doi/pdf/10.1029/2018JE005668}.

\bibitem{Williams2017}
\bibinfo{author}{Williams, J.-P.}, \bibinfo{author}{Paige, D.~A.}, \bibinfo{author}{Greenhagen, B.~T.} \& \bibinfo{author}{Sefton-Nash, E.}
\newblock \bibinfo{journal}{\bibinfo{title}{The global surface temperatures of the {Moon} as measured by the {Diviner Lunar Radiometer Experiment}}}.
\newblock {\emph{\JournalTitle{Icarus}}} \textbf{\bibinfo{volume}{283}}, \bibinfo{pages}{300--325}, \url{10.1016/j.icarus.2016.08.012} (\bibinfo{year}{2017}).
\newblock \bibinfo{note}{Lunar Reconnaissance Orbiter - Part II}.

\bibitem{Williams2019}
\bibinfo{author}{Williams, J.-P.} \emph{et~al.}
\newblock \bibinfo{journal}{\bibinfo{title}{Seasonal polar temperatures on the moon}}.
\newblock {\emph{\JournalTitle{Journal of Geophysical Research: Planets}}} \textbf{\bibinfo{volume}{124}}, \bibinfo{pages}{2505--2521}, \url{https://doi.org/10.1029/2019JE006028} (\bibinfo{year}{2019}).
\newblock \eprint{https://agupubs.onlinelibrary.wiley.com/doi/pdf/10.1029/2019JE006028}.

\bibitem{Schorghofer2020}
\bibinfo{author}{Schorghofer, N.} \& \bibinfo{author}{Williams, J.-P.}
\newblock \bibinfo{journal}{\bibinfo{title}{Mapping of ice storage processes on the moon with time-dependent temperatures}}.
\newblock {\emph{\JournalTitle{The Planetary Science Journal}}} \textbf{\bibinfo{volume}{1}}, \bibinfo{pages}{54}, \url{10.3847/PSJ/abb6ff} (\bibinfo{year}{2020}).

\bibitem{powell2023high}
\bibinfo{author}{Powell, T.} \emph{et~al.}
\newblock \bibinfo{journal}{\bibinfo{title}{High-resolution nighttime temperature and rock abundance mapping of the moon using the diviner lunar radiometer experiment with a model for topographic removal}}.
\newblock {\emph{\JournalTitle{Journal of Geophysical Research: Planets}}} \textbf{\bibinfo{volume}{128}}, \bibinfo{pages}{e2022JE007532} (\bibinfo{year}{2023}).

\bibitem{hayne2017global}
\bibinfo{author}{Hayne, P.~O.} \emph{et~al.}
\newblock \bibinfo{journal}{\bibinfo{title}{Global regolith thermophysical properties of the moon from the diviner lunar radiometer experiment}}.
\newblock {\emph{\JournalTitle{Journal of Geophysical Research: Planets}}} \textbf{\bibinfo{volume}{122}}, \bibinfo{pages}{2371--2400} (\bibinfo{year}{2017}).

\bibitem{lemoine2014grgm900c}
\bibinfo{author}{Lemoine, F.~G.} \emph{et~al.}
\newblock \bibinfo{journal}{\bibinfo{title}{Grgm900c: A degree 900 lunar gravity model from grail primary and extended mission data}}.
\newblock {\emph{\JournalTitle{Geophysical research letters}}} \textbf{\bibinfo{volume}{41}}, \bibinfo{pages}{3382--3389} (\bibinfo{year}{2014}).

\bibitem{Goossens2020}
\bibinfo{author}{Goossens, S.} \emph{et~al.}
\newblock \bibinfo{journal}{\bibinfo{title}{High‐resolution gravity field models from grail data and implications for models of the density structure of the moon's crust}}.
\newblock {\emph{\JournalTitle{Journal of Geophysical Research: Planets}}} \textbf{\bibinfo{volume}{125}}, \bibinfo{pages}{1--31}, \url{10.1029/2019JE006086} (\bibinfo{year}{2020}).

\bibitem{Park2025}
\bibinfo{author}{Park, R.~S.} \emph{et~al.}
\newblock \bibinfo{journal}{\bibinfo{title}{Thermal asymmetry in the moon’s mantle inferred from monthly tidal response}}.
\newblock {\emph{\JournalTitle{Nature}}} \textbf{\bibinfo{volume}{641}}, \bibinfo{pages}{1188--1192}, \url{10.1038/s41586-025-08949-5} (\bibinfo{year}{2025}).

\bibitem{Fortezzo2020}
\bibinfo{author}{{Fortezzo}, C.~M.}, \bibinfo{author}{{Spudis}, P.~D.} \& \bibinfo{author}{{Harrel}, S.~L.}
\newblock \bibinfo{title}{{Release of the Digital Unified Global Geologic Map of the Moon at 1:5,000,000-Scale}}.
\newblock In \emph{\bibinfo{booktitle}{51st Annual Lunar and Planetary Science Conference}}, Lunar and Planetary Science Conference, \bibinfo{pages}{2760} (\bibinfo{year}{2020}).

\bibitem{McClernan2025}
\bibinfo{author}{McClernan, M.~T.} \emph{et~al.}
\newblock \bibinfo{title}{Lunar grid systems, coordinate systems, and map projections for the {Artemis} missions and lunar surface navigation}.
\newblock \bibinfo{type}{Techniques and Methods} \bibinfo{number}{book 11, chap. E1}, \bibinfo{institution}{U.S. Geological Survey} (\bibinfo{year}{2025}).
\newblock \url{10.3133/tm11E1}.

\bibitem{robbins2019new}
\bibinfo{author}{Robbins, S.~J.}
\newblock \bibinfo{journal}{\bibinfo{title}{A new global database of lunar impact craters> 1--2 km: 1. crater locations and sizes, comparisons with published databases, and global analysis}}.
\newblock {\emph{\JournalTitle{Journal of Geophysical Research: Planets}}} \textbf{\bibinfo{volume}{124}}, \bibinfo{pages}{871--892} (\bibinfo{year}{2019}).

\bibitem{Heyer2023}
\bibinfo{author}{{Heyer}, T.} \emph{et~al.}
\newblock \bibinfo{journal}{\bibinfo{title}{{A comparative analysis of global lunar crater catalogs using OpenCraterTool - An open source tool to determine and compare crater size-frequency measurements}}}.
\newblock {\emph{\JournalTitle{\planss}}} \textbf{\bibinfo{volume}{231}}, \bibinfo{pages}{105687}, \url{10.1016/j.pss.2023.105687} (\bibinfo{year}{2023}).

\bibitem{hargitai2025clusters}
\bibinfo{author}{Hargitai, H.} \& \bibinfo{author}{Bro{\v{z}}, P.}
\newblock \bibinfo{journal}{\bibinfo{title}{Clusters of irregular patches on the moon: A new gis-based catalog}}.
\newblock {\emph{\JournalTitle{Icarus}}} \textbf{\bibinfo{volume}{429}}, \bibinfo{pages}{116439} (\bibinfo{year}{2025}).

\bibitem{coyan2025prospectivity}
\bibinfo{author}{Coyan, J.}, \bibinfo{author}{Siegler, M.}, \bibinfo{author}{Martinez-Comacho, J.}, \bibinfo{author}{Beyer, R.} \& \bibinfo{author}{Shirley, M.}
\newblock \bibinfo{journal}{\bibinfo{title}{Prospectivity modeling of the nasa viper landing site at mons mouton near the lunar south pole}}.
\newblock {\emph{\JournalTitle{The Planetary Science Journal}}} \textbf{\bibinfo{volume}{6}}, \bibinfo{pages}{105} (\bibinfo{year}{2025}).

\bibitem{he2016deep}
\bibinfo{author}{He, K.}, \bibinfo{author}{Zhang, X.}, \bibinfo{author}{Ren, S.} \& \bibinfo{author}{Sun, J.}
\newblock \bibinfo{title}{Deep residual learning for image recognition}.
\newblock In \emph{\bibinfo{booktitle}{Proceedings of the IEEE conference on computer vision and pattern recognition}}, \bibinfo{pages}{770--778} (\bibinfo{year}{2016}).

\bibitem{liu2022swin}
\bibinfo{author}{Liu, Z.} \emph{et~al.}
\newblock \bibinfo{title}{Swin transformer v2: Scaling up capacity and resolution}.
\newblock In \emph{\bibinfo{booktitle}{Proceedings of the IEEE/CVF conference on computer vision and pattern recognition}}, \bibinfo{pages}{12009--12019} (\bibinfo{year}{2022}).

\bibitem{gomes2025terratorch}
\bibinfo{author}{Gomes, C.}, \bibinfo{author}{Blumenstiel, B.} \emph{et~al.}
\newblock \bibinfo{journal}{\bibinfo{title}{Terratorch: The geospatial foundation models toolkit}}.
\newblock {\emph{\JournalTitle{arXiv preprint arXiv:2503.20563}}}  (\bibinfo{year}{2025}).

\bibitem{ren2015faster}
\bibinfo{author}{Ren, S.}, \bibinfo{author}{He, K.}, \bibinfo{author}{Girshick, R.} \& \bibinfo{author}{Sun, J.}
\newblock \bibinfo{journal}{\bibinfo{title}{Faster r-cnn: Towards real-time object detection with region proposal networks}}.
\newblock {\emph{\JournalTitle{Advances in neural information processing systems}}} \textbf{\bibinfo{volume}{28}} (\bibinfo{year}{2015}).

\bibitem{ronneberger2015u}
\bibinfo{author}{Ronneberger, O.}, \bibinfo{author}{Fischer, P.} \& \bibinfo{author}{Brox, T.}
\newblock \bibinfo{title}{U-net: Convolutional networks for biomedical image segmentation}.
\newblock In \emph{\bibinfo{booktitle}{International conference on medical image computing and computer-assisted intervention}}, \bibinfo{pages}{234--241} (\bibinfo{organization}{Springer}, \bibinfo{year}{2015}).

\bibitem{wagner2024crater}
\bibinfo{author}{Wagner, R.~V.} \emph{et~al.}
\newblock \bibinfo{journal}{\bibinfo{title}{Where is that crater? best practices for obtaining accurate coordinates from lroc nac data}}.
\newblock {\emph{\JournalTitle{The Planetary Science Journal}}} \textbf{\bibinfo{volume}{5}}, \bibinfo{pages}{157} (\bibinfo{year}{2024}).

\end{thebibliography}

\clearpage

\end{document}